\documentclass[lettersize,journal]{IEEEtran}
\usepackage{amsmath,amsfonts}
\usepackage[linesnumbered,ruled]{algorithm2e}
\usepackage{array}
\usepackage[caption=false,font=normalsize,labelfont=sf,textfont=sf]{subfig}
\usepackage{textcomp}
\usepackage{stfloats}
\usepackage{url}
\usepackage{verbatim}
\usepackage{graphicx}
\usepackage{cite}
\usepackage{amsthm}
\usepackage{amsfonts}
\usepackage{multirow}
\usepackage{bm}
\usepackage{rotating}
\usepackage{pifont}
\usepackage{makecell}
\usepackage{booktabs}
\newcommand{\figref}[1]{Figure \ref{#1}}
\newcommand{\tabref}[1]{Table \ref{#1}}

\newcommand{\equref}[1]{Equation (\ref{#1})}
\newcommand{\appendixref}[1]{Appendix}
\newcommand{\algoref}[1]{Algorithm \ref{#1}}

\begin{document}

\title{A Simple Framework for Multi-mode Spatial-Temporal Data Modeling}

\author{Zihang~Liu,
        Le~Yu,
        Tongyu~Zhu,
        Leiei~Sun
\IEEEcompsocitemizethanks{\IEEEcompsocthanksitem Z. Liu, L. Yu, T. Zhu and L. Sun are with the State Key Laboratory of Software Development Environment, Beihang University, Beijing, 100191, China.\protect\\
E-mail: \{lzhmark,yule,zhutongyu,leileisun\}@buaa.edu.cn
}
\thanks{Manuscript received August 22, 2023.}}

\markboth{Journal of \LaTeX\ Class Files,~Vol.~14, No.~8, August~2021}%
{Shell \MakeLowercase{\textit{et al.}}: A Sample Article Using IEEEtran.cls for IEEE Journals}

\IEEEpubid{0000--0000/00\$00.00~\copyright~2021 IEEE}

\maketitle

\begin{abstract}
Spatial-temporal data modeling aims to mine the underlying spatial relationships and temporal dependencies of objects in a system.
However, most existing methods focus on the modeling of spatial-temporal data in a single mode, lacking the understanding of multiple modes. Though very few methods have been presented to learn the multi-mode relationships recently, they are built on complicated components with higher model complexities.
In this paper, we propose a simple framework for multi-mode spatial-temporal data modeling to bring both effectiveness and efficiency together. Specifically, we design a general cross-mode spatial relationships learning component to adaptively establish connections between multiple modes and propagate information along the learned connections. Moreover, we employ multi-layer perceptrons to capture the temporal dependencies and channel correlations, which are conceptually and technically succinct.
Experiments on three real-world datasets show that our model can consistently outperform the baselines with lower space and time complexity, opening up a promising direction for modeling spatial-temporal data. The generalizability of the cross-mode spatial relationships learning module is also validated. Codes and datasets are available at \url{https://github.com/lzhmarkk/SimMST}.
\end{abstract}


\section{Introduction}


Spatial-temporal data record the information of objects across both space and time in a system, which are pervasive in many applications, such as traffic forecasting 
\cite{Li_DCRNN_Diffusion_2018, Yu_STGCN_SpatioTemporal_2018}, air quality inference \cite{DBLP:conf/kdd/YiZWLZ18, wen2019novel}, and electricity consumption prediction \cite{Wu_MTGNN_Connecting_2020, Ye_ESG_Learning_2022}. In recent years, spatial-temporal modeling has attracted a lot of attention, whose core is to mine the underlying spatial relationships and temporal dependencies of objects \cite{Wu_GraphWaveNet_Graph_2019, DBLP:journals/tkde/WangCY22}.

\begin{figure}[!htbp]
    \centering
    \includegraphics[width = \columnwidth]{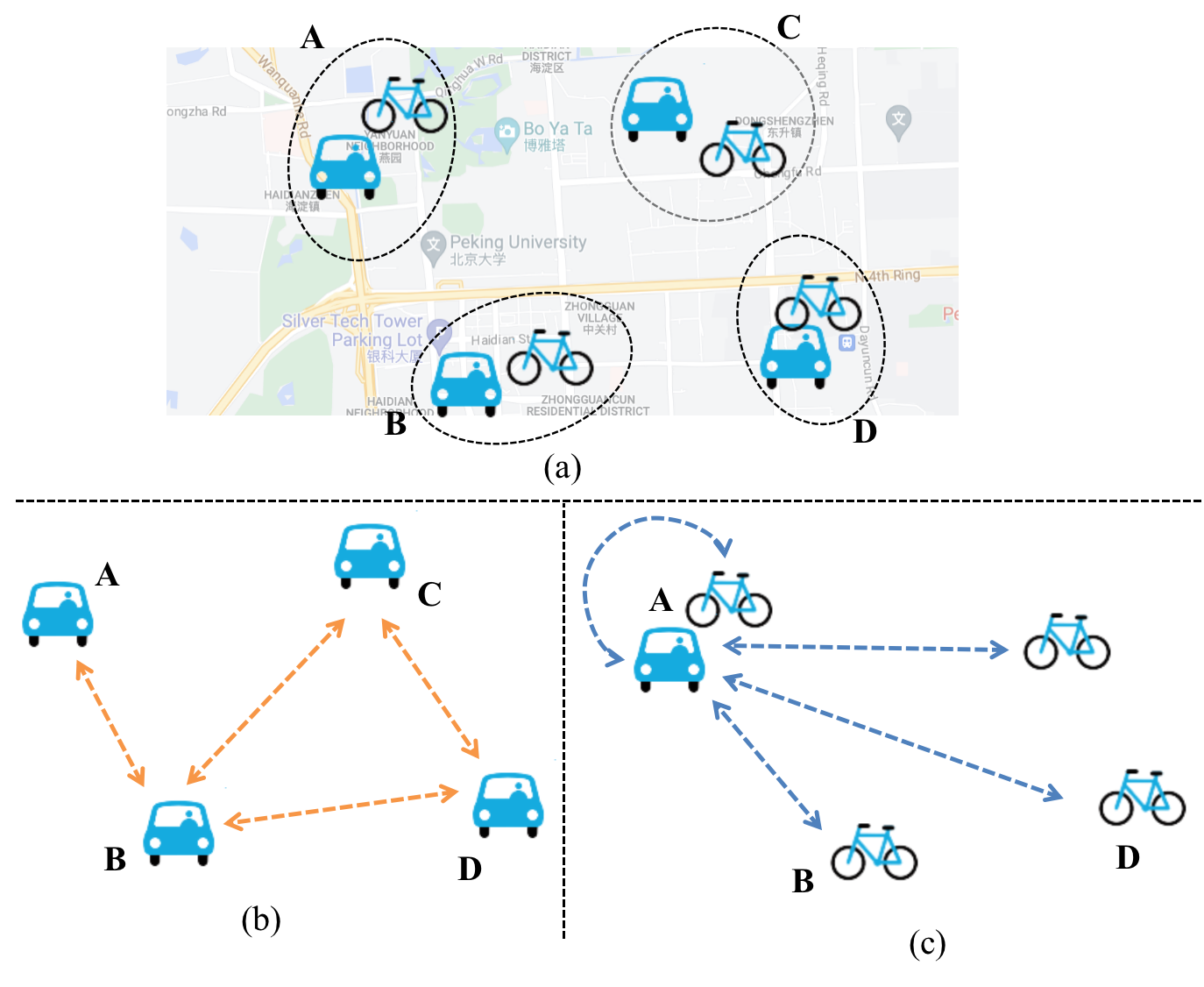}
    \caption{Illustrations of spatial and mode dependency of multi-mode spatial-temporal data. (a) Multiple modes of a specific location can be simultaneously observed in many real-world scenarios, e.g. bike and taxi demands of each location. (b) Most previous works consider the spatial relationships between locations with only a single mode. (c) The correlations between different modes remain to be investigated.}
    \label{fig:Illustration}
\end{figure}

Existing methods for spatial-temporal data modeling mainly learn from a single mode \cite{Li_DCRNN_Diffusion_2018,Yu_STGCN_SpatioTemporal_2018,DBLP:conf/kdd/YiZWLZ18,Wu_GraphWaveNet_Graph_2019,Guo_ASTGCN_Attention_2019,wen2019novel,DBLP:conf/aaai/ChenCXCGF20,DBLP:conf/aaai/ZhengFW020,Wu_MTGNN_Connecting_2020,Oreshkin_FC-GAGA_FCGAGA_2021,Ye_ESG_Learning_2022,Li_MTS-Mixers_MTSMixers_2023}. For example, \cite{Yu_STGCN_SpatioTemporal_2018} combined the Graph Neural Networks (GNNs) \cite{DBLP:conf/nips/DefferrardBV16} and convolutional sequence learning modules to jointly extract spatial-temporal features for traffic speed forecasting. \cite{Wu_MTGNN_Connecting_2020} designed a GNN-based framework to exploit the inherent relationships of different single-mode sequences. \cite{Oreshkin_FC-GAGA_FCGAGA_2021} extended \cite{Oreshkin_N-BEATS_NBEATS_2020} with a graph gate for spatial modeling and a time gate to capture multiplicative seasonality.
However, in real-world scenarios, multiple factors of a specific location can be simultaneously observed (i.e., multi-mode) and tend to affect each other. For example, the demands of taxis and bikes both exist in one certain area and the changes in taxi demands are often influenced by the variations in bike demands \cite{Ye_CoST-Net_CoPrediction_2019,Liu_CoGNN_CoPrediction_2021}. Therefore, it is essential to investigate the relationships between multiple modes, which cannot be captured by previous methods. \figref{fig:Illustration} demonstrates the spatial and mode dependency of multi-mode spatial-temporal data.

\begin{table*}[!htbp]
    \centering
    \normalsize
    \caption{Comparison of components, time complexity and space complexity of single-mode and multi-mode methods. $N$, $T$, $L$, $C$ and $K$ respectively refers to the number of objects, timestamps, layers, features and GCN hops. \textit{Other} stands for other components such as embedding layers and normalization layers. For fair comparison, we ignore the number of modes $M$ here.}
    \label{tab:ComponentsCompare}
    \resizebox{2.0\columnwidth}{!}{
        \setlength{\tabcolsep}{1.5mm}{
            \begin{tabular}{cccccccccccc}
            \toprule
            \multirow{3}{*}{Methods} & \multirow{3}{*}{\makecell{Multi\\mode}} & \multicolumn{5}{c}{Spatial} & \multicolumn{3}{c}{\multirow{2}{*}{Temporal}} & \multicolumn{2}{c}{\multirow{2}{*}{Other}}\\
            & & \multicolumn{2}{c}{Learning} & \multicolumn{3}{c}{Propagation} & & & & & \\
            \cmidrule(r){3-4} \cmidrule(r){5-7} \cmidrule(r){8-10} \cmidrule(r){11-12} 
            & & Time & Space & Component & Time & Space & Component & Time & Space & Time & Space \\
            \midrule
            STGCN & $\times$ & - & - & GCN & $\mathcal{O}\left(N^2TLK\right)$ & $\mathcal{O}\left(LK\right)$ & TCN & $\mathcal{O}\left(NTL\right)$ & $\mathcal{O}\left(L\right)$ & $\mathcal{O}\left(NT(L+C)\right)$ & $\mathcal{O}\left(NL+C\right)$ \\
            GWNet & $\times$ & $\mathcal{O}\left(N^2\right)$ & $\mathcal{O}\left(N\right)$ & GCN & $\mathcal{O}\left(N^2TLK\right)$ & $\mathcal{O}\left(LK\right)$ & TCN & $\mathcal{O}\left(NTL\right)$ & $\mathcal{O}\left(L\right)$ & $\mathcal{O}\left(NT(L+C)\right)$ & $\mathcal{O}\left(L+C\right)$\\
            MTGNN & $\times$ & $\mathcal{O}\left(N^2\right)$ & $\mathcal{O}\left(N\right)$ & GCN & $\mathcal{O}\left(N^2TLK\right)$ & $\mathcal{O}\left(LK\right)$ & TCN & $\mathcal{O}\left(NTL\right)$ & $\mathcal{O}\left(L\right)$ & $\mathcal{O}\left(NT(L+C)\right)$ & $\mathcal{O}\left(NTL+C\right)$ \\
            ESG & $\times$ & $\mathcal{O}\left(N^2T\right)$ & $\mathcal{O}\left(N+T\right)$ & GCN & $\mathcal{O}\left(N^2TLK\right)$ & $\mathcal{O}\left(LK\right)$ & TCN & $\mathcal{O}\left(NTL\right)$ & $\mathcal{O}\left(L\right)$ & $\mathcal{O}\left(NT(L+C)\right)$ &
            $\mathcal{O}\left(NTL+C\right)$ \\
            FC-GAGA & $\times$ & $\mathcal{O}\left(N^2\right)$ & $\mathcal{O}\left(N\right)$ & MLP & $\mathcal{O}\left(N^2TL\right)$ & - & MLP & $\mathcal{O}\left(N^2TL\right)$ & $\mathcal{O}\left(NTL\right)$ & 
            $\mathcal{O}\left(NT(L+C)\right)$ & 
            $\mathcal{O}\left(TL+C\right)$ \\
            MTS-Mixer & $\times$ & $\mathcal{O}\left(N^2\right)$ & $\mathcal{O}\left(N^2\right)$ & MLP & $\mathcal{O}\left(N^2TC\right)$ & - & MLP & $\mathcal{O}\left(NT^2C\right)$ & $\mathcal{O}\left(T^2\right)$ & $\mathcal{O}\left(NTLC\right)$ & $\mathcal{O}\left(NL\right)$ \\
            \cmidrule(r){1-12}
            MOHER & \checkmark & - & - & GCN & $\mathcal{O}\left(N^2TK\right)$ & $\mathcal{O}\left(K\right)$ & LSTM & $\mathcal{O}\left(NT\right)$ & $\mathcal{O}\left(1\right)$ & $\mathcal{O}\left(NT(L+C)\right)$ & $\mathcal{O}\left(C\right)$ \\
            CoGNN & \checkmark & $\mathcal{O}\left(N^2\right)$ & $\mathcal{O}\left(N\right)$ & GCN & $\mathcal{O}\left(N^2TLK\right)$ & $\mathcal{O}\left(LK\right)$ & TCN & $\mathcal{O}\left(NTL\right)$ & $\mathcal{O}\left(L\right)$ & 
            $\mathcal{O}\left(NT(L+C)\right)$ & 
            $\mathcal{O}\left(NTL+C\right)$ \\
            SimMST & \checkmark & $\mathcal{O}\left(N^2\right)$ & $\mathcal{O}\left(N\right)$ & MLP & $\mathcal{O}\left(N^2T\right)$ & - & MLP & $\mathcal{O}\left(NT^2\right)$ & $\mathcal{O}\left(T^2\right)$ & 
            $\mathcal{O}\left(NTC\right)$ & 
            $\mathcal{O}\left(T+C\right)$ \\
            \bottomrule
            \end{tabular}
        }
    }
\end{table*}

In recent years, very few efforts have been made on modeling multi-mode spatial-temporal data. One part of the methods restricted their studied objects to be organized in regular grids in the space, and thus reduced its usability in scenarios with irregular regions \cite{Liang_GeoMAN_GeoMAN_2018,Huang_MiST_MiST_2019,Ye_CoST-Net_CoPrediction_2019,Deng_CEST_Pulse_2021}. Other methods tried to apply GNNs to break the constraints of grids. For instance, 
MOHER \cite{Zhou_MOHER_Modeling_2021} and CoGNN \cite{Liu_CoGNN_CoPrediction_2021} utilized GNNs to propagate information along the connections between nodes and modes to explore multiple modes' relationships.
Although insightful, they are built on complicated modules, which inevitably introduce more learnable parameters and incur higher computational costs.
\IEEEpubidadjcol
We provide the analysis of existing methods in \tabref{tab:ComponentsCompare}, which shows that they either fail to capture multi-mode relationships or require complex components.
Specifically, most methods use Temporal Convolutional Networks (TCNs) to learn temporal dependencies. They concatenate multiple filters with different kernel sizes to capture temporal patterns with different granularities and use dilation configurations for larger receptive fields. Moreover, these methods stack multiple GNN layers to propagate high-order spatial information. ESG \cite{Ye_ESG_Learning_2022} even entangles recurrent modules with graph structure learning. These complex designs are computationally expensive but might not be necessary.

To this end, we aim to investigate \textit{whether it is necessary to design sophisticated components for modeling multi-mode spatial-temporal data}. To be specific, we propose a \textbf{Sim}ple framework for \textbf{M}ulti-mode \textbf{S}patial-\textbf{T}emporal data modeling (SimMST) to bring both effectiveness and efficiency together. A general cross-mode spatial relationships learning component is designed, which can adaptively establish connections between multiple modes and propagate information along the learned connections. Also, we leverage multi-layer perceptrons to capture the temporal dependencies and channel correlations. Our approach is built by stacking the above components multiple times and can be trained in an end-to-end manner. We conduct extensive experiments on three real-world datasets, and the results demonstrate that our model outperforms existing methods with fewer learnable parameters and lower computational costs. We also integrate the proposed cross-mode spatial relationships learning module into existing methods and find that this module is general and can bring improvements due to the learning of multi-mode relationships. Our main contributions include:

\begin{itemize}
    \item We are the first to show the task of multi-mode spatial-temporal data modeling can be well tackled by a simple framework, which can simultaneously learn the temporal dependencies, cross-mode spatial relationships, and channel correlations.
    
    \item A cross-mode spatial relationships learning component is designed, which can adaptively learn connections between different modes and propagate information along the connections. This module is general and applicable to existing spatial-temporal modeling methods.
    
    \item Extensive experimental results show the superiority of our approach over existing methods, which motivates future research to rethink the key factors for modeling multi-mode spatial-temporal data as well as the potential of simple architectures.
\end{itemize}

\section{Preliminaries}

\begin{table}[!htbp]
    \centering
    \normalsize
    \caption{Some important notations used in this paper.}
    \label{tab:notations}
    \resizebox{0.9\columnwidth}{!}{
        \setlength{\tabcolsep}{1.5mm}{
            \begin{tabular}{c|c}
                \hline
                Notation & Implication \\
                \hline
                $\mathbb{M}$ & The set of all modes (e.g. railway, bus) \\
                $\mathbb{N}$ & The set of all objects (e.g. regions) \\
                \hline
                $M$ & Total number of modes \\
                $N$ & Total number of objects \\
                $T$ & Total number of timestamps \\
                $C$ & Total number of features \\
                $L$ & Total number of model layers \\
                \hline
                $m$ & A mode\\
                $n$ & An object \\
                $t$ & A timestamp \\
                $c$ & A feature \\
                \hline
                $\mathbf{A}_{m_i,m_j}$ & Relationship matrix between modes $m_i$ and $m_j$ \\
                $\mathbf{H^l}$ & Hidden state of layer $l$ \\
                $\mathbf{X}$ & Input history observations \\
                $\mathbf{\hat{Y}}$ & Output predicted observations \\
                $\mathbf{X}_m$ & Input history observations with mode $m$ \\
                $\mathbf{\hat{Y}}_m$ & Output predicted observations with mode $m$ \\
                \hline
            \end{tabular}
        }
    }
\end{table}


\subsection{Multi-mode Spatial-Temporal Data}
Let $\mathbb{N}=\{n_1,\cdots,n_N\}$ denote the set of all the studied objects with $N$ as the cardinality. Let $\mathbb{M}=\{m_1,\cdots,m_M\}$ represents the set of all the modes with $M$ as the cardinality. We denote observations of object $n \in \mathbb{N}$ with mode $m \in \mathbb{M}$ at time $t$ by $\bm{X}_{m,n,t,*}\in\mathbb{R}^C$, where $C$ is the feature dimension and $*$ refers to all entities in the corresponding dimension. Therefore, the observations of all objects with mode $m$ at all timestamps can be denoted as $\bm{X}_{m,*,*,*}\in\mathbb{R}^{N\times T\times C}$, where $T$ is the total number of timestamps. By stacking $M$ modes, we can represent the multi-mode spatial-temporal data as $\bm{X}\in\mathbb{R}^{M\times N\times T\times C}$. Take the traffic forecasting task as an example, each region has observations with several traffic modes at different times, such as the demand of taxis, bikes, and buses, which could be formalized as multi-mode spatial-temporal data.

\subsection{Multi-mode Relation Matrices}
A multi-mode relation matrix captures the relationships between different objects with two specific modes. Formally, we use matrix $\bm{A}_{m_i,m_j}\in \mathbb{R}^{N\times N}$ to record the relationships of $N$ objects with modes $m_i, m_j \in \mathbb{M}$. Therefore, for $M$ modes, we have $M\times M$ matrices that represent the pairwise relationships between all modes, that is, $\mathcal{A}=\{\bm{A}_{m_i,m_j} | m_i,m_j \in \mathbb{M}\}$.

\subsection{Problem Formulation}
In this paper, we formalize the multi-mode spatial-temporal modeling problem as a forecasting task. Given the observations within historical $W$ timestamps $\bm{X}_{*,*,t-W+1:t,*} \in \mathbb{R}^{M\times N \times W \times C}$, we aim to design an algorithm $\mathcal{F}$ to predict future observations in the next $H$ horizons for all the objects with all the modes, i.e., $\hat{\bm{Y}}_{*,*,t+1:t+H,*} \in \mathbb{R}^{M \times N \times H \times C}$, which can be denoted by the following equation,
\begin{equation}
    \label{equ:Formulation}
    \hat{\bm{Y}}_{*,*,t+1:t+H,*}=\mathcal{F}\left(\bm{X}_{*,*,t-W+1:t,*}\right).
\end{equation}




\tabref{tab:notations} provides an overview of some important notations used in this paper.

\section{Methodology}

\begin{figure*}[!t]
    \centering
    \includegraphics[width = 1.68\columnwidth]{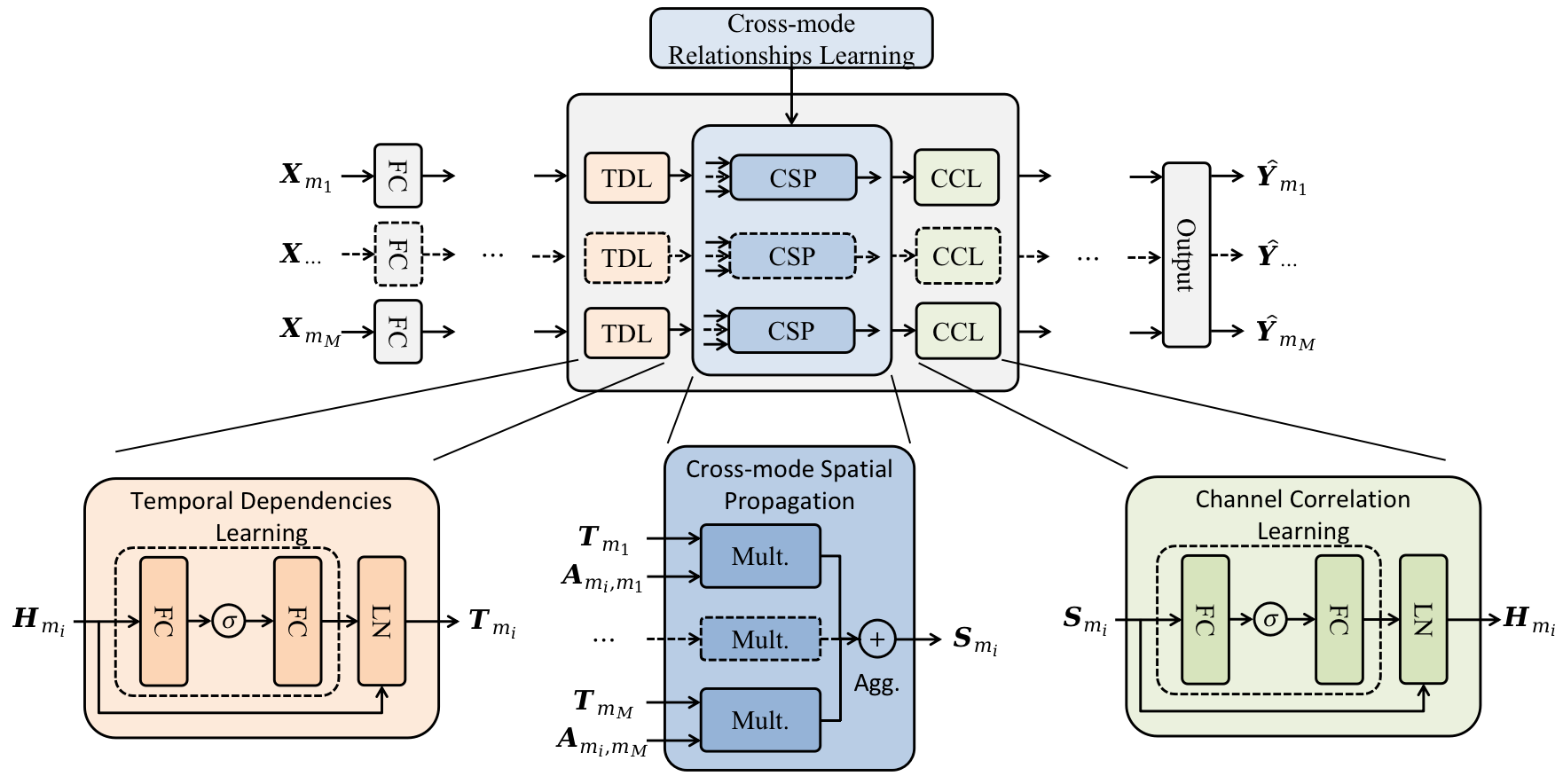} 
    \caption{Framework of the proposed SimMST.}
    \label{fig:Framework}
\end{figure*}

In this section, we elaborate on the design of the overall framework and each component.

\subsection{Overview}

Figure \ref{fig:Framework} presents the framework of our method, which learns from the perspectives of temporal, spatial, cross-mode, and channel. The multi-mode spatial-temporal input $\{\bm{X}_{m_1},\bm{X}_{m_2},\cdots,\bm{X}_{m_M}\}$ is first fed into multiple mode-specific embedding layers $\{f_1^{init},f_2^{init},\cdots,f_M^{init}\}$ to obtain the corresponding initial hidden states $\{\bm{H}_{m_1}^0, \bm{H}_{m_2}^0, \cdots, \bm{H}_{m_M}^0\}$.
Then a Temporal Dependencies Learning (TDL) component is utilized to model temporal evolutionary patterns for each mode, and a Cross-mode Spatial Relationships Learning (CSRL) component is used to capture both spatial relationships and interactions between different modes. Third, a Channel Correlations Learning (CCL) component is employed to learn the correlations of different channels. We stack the above components into layers to obtain hierarchical hidden states and send all hidden states to an output module to generate predictions with each mode.

\subsection{Temporal Dependencies Learning}
\label{section:TDL}


Given hidden states of $l-1$ layer $\bm{H}_m^{l-1}\in \mathbb{R}^{N\times\ T_{l-1}\times D}$ with time length $T_{l-1}$, \textbf{T}emporal \textbf{D}ependencies \textbf{L}earning (TDL) component models the temporal evolutionary patterns of each object within each mode, which reflect the inherent patterns of each object.

TDL regards the input as time series and conducts information mixing with neural networks $f^{TDL}$ along the temporal dimension. We use a layer normalization operation for regularization and a residual connection to retain the input features.
The process of TDL can be denoted as:
\begin{equation}
    \label{equ:TDL}
    \bm{T}_m^l=LN(f^{TDL}_m(\bm{H}_m^{l-1}))+\bm{H}_m^{l-1},
\end{equation}

\noindent where $f^{TDL}_m$ denotes the neural networks applied on the temporal dimension and $m$ refers to mode.

We proposed several forms of empirical configurations of TDL:

1) \textit{General configuration}. Multi-layer perceptron (MLP) learns the pair-wise influence from each input timestamp to each output timestamp. It has a full receptive field and is naturally sensitive to the input order. Therefore it is suitable to deal with sequences. The length of temporal dimension is reduced by half, which helps to learn high-level temporal information. It can be described with the following equation:
\begin{equation}
    \label{equ:TDL_MLP}
    f^{TDL}_m(\bm{H})=\sigma(\bm{H}\bm{W}_{m,0}^{t,l}+\bm{b}_{m,0}^{t,l})\bm{W}_{m,1}^{t,l}+\bm{b}_{m,1}^{t,l},
\end{equation}

\noindent where the activation function $\sigma(\cdot)$ is GeLU.
$\bm{W}_{m}^{t,l}$ and $\bm{b}_{m}^{t,l}$ are learnable temporal weight and bias parameters.

2) \textit{Seasonality configuration}. Seasonality describes the recurring and regular characteristics of a series. We adopt Fourier Transforms to discriminate the input into a series of trigonometric functions to better capture the seasonal patterns. It can be described with the following equation:
\begin{equation}
    \label{equ:TDL_FFT}
    f^{TDL}_m(\bm{H})=\mathcal{F}^{-1}\left(\mathcal{F}(\bm{H})\odot\bm{W}_{m}^{t,l}+\bm{b}_{m}^{t,l}\right),
\end{equation}

\noindent where $\mathcal{F}$ represents FFT operation and $\mathcal{F}^{-1}$ represents inverse FFT operation. $\odot$ refers to Hadamard product.

In this paper, we choose the MLPs as the backbone of TDL. Each timestamp of the output matrix $\bm{T}_m^l$ receives direct information from all timestamps of the input, and thus the interactions of different timestamps can be learned.

\subsection{Cross-mode Spatial Relationships Learning}

In this component, we aim to model the spatial and multi-mode relationships, which correspond to the $N$ dimension and $M$ dimension of input $\bm{X}$ respectively. The spatial relationships indicate the relations between objects within a mode. For example, the railway demands of an area can be influenced by the railway demands of another area.
The multi-mode relationships indicate the impacts between objects with different modes, which are pervasive in real-world scenarios. Intuitively, the railway demand in an area is positively correlated with taxis (the demand brought by the railways will be absorbed by taxis), but the demand for bikes is inversely related to taxis (people either choose bikes or taxis for travel).

In order to explicitly model both spatial and multi-mode relationships, we propose a \textbf{C}ross-mode \textbf{S}patial \textbf{R}elationships \textbf{L}earning (CSRL) component. It consists of a cross-mode relationships learning module and several cross-mode spatial propagation modules.

\textbf{Cross-mode Relationships Learning Module.} For every two modes $m_i$ and $m_j$, we use a relation matrix $\bm{A}_{m_i,m_j}$ to record the learned connections from mode $m_j$ to mode $m_i$.
First, we denote the trainable object embeddings with mode $m$ as $\bm{E}_m\in\mathbb{R}^{N\times D_{emb}}$, where $D_{emb}$ is the embedding dimension. Then, two MLPs are respectively utilized to retrieve the in-affect embeddings $\bm{E}_m^{in}$ and out-affect embeddings $\bm{E}_m^{out}$. The relation matrix between mode $m_j$ and $m_i$ can be denoted as:
\begin{equation}
    \label{equ:SpatialLearning}
    \Tilde{\bm{A}}_{m_i,m_j} = \psi\left(\tau(\bm{E}_{m_j}^{out}\times(\bm{E}_{m_i}^{in})^T-\bm{E}_{m_j}^{in}\times(\bm{E}_{m_i}^{out})^T)\right),
\end{equation}
\noindent where $(\cdot)^T$ refers to matrix transposition.
The former term in Equation \ref{equ:SpatialLearning} represents the gross impact from mode $m_j$ to $m_i$, and the latter represents the reverse impact. We can obtain the net impact from mode $m_j$ to $m_i$ with a subtraction. $\psi(\cdot)$ and $\tau(\cdot)$ are respectively ReLU and Tanh activation functions.
We further facilitate the training with self-loops and normalization:
\begin{equation}
    \label{equ:SpatialLearningNorm}
    \bm{A}_{m_i,m_j} = \bm{D}_{m_i,m_j}^{-1}(\Tilde{\bm{A}}_{m_i,m_j} + \bm{I}),
\end{equation}
\noindent where $\bm{D}_{m_i,m_j}$ is the degree matrix of $\Tilde{\bm{A}}_{m_i,m_j} + \bm{I}$.

\textbf{Cross-mode Spatial Propagation Module.} Then, we propose a Cross-mode Spatial Propagation module to simultaneously model the spatial and multi-mode relationships, which intakes multi-mode embeddings $\{\bm{T}_m^l|m\in\mathbb{M}\}$ output from previous TDL component and relation matrices $\{\bm{A}_{m_i,m_j} |  m_i,m_j\in\mathbb{M}\}$ acquired from above relationships learning module.

Taking mode $m_i$ as a target, we use matrix multiplication to propagate information along the connections in $\bm{A}_{m_i,*}$ from two aspects to compute the impacts from each source mode $m_j\in \mathbb{M}$:
\begin{equation}
    \label{equ:SpatialPropagation}
    \hat{\bm{S}}_{m_i,m_j}^l=\bm{A}_{m_i,m_j}\bm{T}_{m_j}^l+(\bm{A}_{m_i,m_j})^T\bm{T}_{m_j}^l.
\end{equation}

Then, we aggregate the impacts from all source modes with a weighted summation:
\begin{equation}
    \label{equ:SpatialAggregation}
    \bm{S}_{m_i}^l=\sum_{m_j\in\mathbb{M}}{\left(w_{m_i,m_j}+\mathbb{I}(i,j)\right)\hat{\bm{S}}_{m_i,m_j}^l},
\end{equation}

\noindent where $\mathbb{I}(i,j)$ is an indication function returning 1 if $i$ equals to $j$, and 0 otherwise. $w_{m_i,m_j}$ is a learnable weight scalar.

The cross-mode spatial relationships learning component proposed in this section is general. It can be easily integrated with existing spatial-temporal modeling methods to empower them to model multi-mode relationships, which can be testified by the experiments in Section \ref{section:Multi-mode}.

\subsection{Channel Correlations Learning}
\label{section:CCL}

In the above sections, the multi-mode, spatial and temporal semantics have been modeled, which respectively correspond to dimensions $M$, $N$, and $T$ of the input with shape $\mathbb{R}^{M\times N \times T \times C}$.
Besides, the correlation of different entries of feature vectors (named \textit{channel} and corresponding to dimension $C$) might contain semantics about the relationship between features. For example, the pick-up demands and drop-off demands of a region are correlated in some way. Explicitly learning the representation of features and their interactions are crucial for multi-mode spatial-temporal modeling.

To this end, we propose \textbf{C}hannel \textbf{C}orrelations \textbf{L}earning (CCL) component to learn the correlation between channels. Formally, given $\bm{S}_m^l$ acquired from \equref{equ:SpatialAggregation}, we use the following process to model channel interactions:
\begin{equation}
    \label{equ:CCL}
    \bm{H}_m^l = LN(f^{CCL}_m(\bm{S}_m^l))+\bm{S}_m^l,
\end{equation}

We adopt a general configuration similar to \equref{equ:TDL_MLP} for the implementation of $f^{CCL}$. Specifically, we use a 2-layer MLP as the backbone to capture the feature interactions and a GeLU activation between the layers to learn non-linear patterns. 


\begin{algorithm}[!htbp]
\SetKwInOut{Input}{Input}
\SetKwInOut{Output}{Output}
\SetKwComment{Comment}{/* }{ */}
    \caption{The process of SimMST.}
    \label{alg:training}
    \Input{Historical observations $\mathbf{X}$ with multiple modes $\mathbb{M}=\{m_1,m_2,\cdots,m_M\}$}
    \Output{Predicted observations $\hat{\mathbf{Y}}$}
        $\{\bm{A}_{m_i,m_j} | m_i,m_j\in\mathbb{M}\}\gets$ Relation matrices learning\;
        $\mathbf{X}_{m_1},\mathbf{X}_{m_2},\cdots,\mathbf{X}_{m_M} \gets$ Split $\mathbf{X}$ by mode\;
        \For{$m\in \mathbb{M}$}{
            $\bm{H}_m^0 \gets f^{init}_m(\bm{X}_m)$
        }
        \For{layer $l\in [1,L]$}{
            \Comment{TDL, \equref{equ:TDL}}
            \For{$m\in \mathbb{M}$}{
                $\bm{T}_m^l \gets$ Learn temporal dependencies\;
            }
            \Comment{CSRL, \equref{equ:SpatialPropagation} and \equref{equ:SpatialAggregation}}
            \For{$m_i,m_j\in \mathbb{M}$}{
                $\hat{\bm{S}}_{m_i,m_j}^l \gets$ Compute impacts from each source mode\;
                $\bm{S}_{m_i}^l \gets$ Aggregate impacts from all source modes\;
            }
            \Comment{CCL, \equref{equ:CCL}}
            \For{$m\in \mathbb{M}$}{
                $\bm{H}_m^l \gets$ Learn channel correlations\;
            }
        }
        \Comment{Output, \equref{equ:BlockAggregation}}
        $\bm{E}^{semantic} \gets$ Encode time semantics\;
        \For{$m\in \mathbb{M}$}{
            $\bm{Z}_m \gets$ Aggregate hidden states and time semantics\;
            $\hat{\bm{Y}}_m \gets f_m^{out}(\bm{Z}_m)$\;
        }
        $\hat{\bm{Y}} \gets$ Concatenate predictions of all modes\;
\end{algorithm}

\subsection{Output Module}

We combine the above TDL, CSRL, and CCL components into a layer and stack multiple layers to obtain hierarchical hidden states $\{\bm{H}_m^l | m\in\mathbb{M}, 0\leq l\leq L\}$ for all modes.
Then, we aggregate the hidden states of each mode with a summation to obtain the summary of input multi-mode spatial-temporal data.

To better model periodical patterns, we extract the timestamp semantics, including: 1) which time slot it belongs to in a day (48 slots per day), and 2) which day it belongs to in a week (7 slots per week) and respectively retrieve the corresponding embedding vectors from a $\mathbb{R}^{48\times C}$ trainable embedding table and a $\mathbb{R}^{7\times C}$ trainable embedding table.

Then, we concatenate the time semantic embedding $\bm{E}^{semantic}$ and the aggregated hidden state:
\begin{equation}
    \label{equ:BlockAggregation}
    \bm{Z}_m = \sum_{l=0}^L{\bm{H}_m^l} || \bm{E}^{semantic}.
\end{equation}

Finally, for each mode $m$, a two-layer MLP $f_m^{out}(\cdot)$ is used to predict future observations in the next $H$ horizons for all the objects:
\begin{equation}
    \label{equ:Output}
    \hat{\bm{Y}}_m = f_m^{out}(\bm{Z}_m) \in \mathbb{R}^{N\times H\times C}.
\end{equation}

\noindent The output of SimMST with all modes are $\{\hat{\bm{Y}}_{m_1},\hat{\bm{Y}}_{m_2},\cdots,\hat{\bm{Y}}_{m_M}\}$. We concatenate them together to obtain the final predictions $\hat{\bm{Y}}$.


\subsection{Model Learning Process}

Our SimMST is trained in an end-to-end manner. Mean Absolute Error (MAE) is adopted to calculate the divergence between the predicted values $\hat{\bm{Y}}_m$ and the ground truth observations $\bm{Y}_m$ for each mode $m$. We use an unweighted summation over modes to simultaneously optimize parameters with all modes:
\begin{equation}
    \label{equ:Optimization}
    \mathcal{L} = \sum_{m\in\mathbb{M}} \sum_{n\in\mathbb{N}} \sum_{t=0}^{H} \sum_{c=0}^{C} |\hat{\bm{Y}}_{m,n,t,c} - \bm{Y}_{m,n,t,c}|.
\end{equation}

The overall process of SimMST is shown in \algoref{alg:training}.

\subsection{Complexity Analysis}

The overall time complexity of SimMST is $\mathcal{O}\left(MNT(C+MN+T)\right)$ and the space complexity is $\mathcal{O}\left(M(C+MN+T^2)\right)$.

\begin{itemize}
    \item In cross-model relationship learning module (\equref{equ:SpatialLearning}), the time and space complexities for all $M^2$ pair of modes are respectively $\mathcal{O}\left(M^2N^2\right)$ and $\mathcal{O}\left(M^2N\right)$. In propagation module (\equref{equ:SpatialPropagation}), the time complexity is $\mathcal{O}\left(M^2N^2T_{l}\right)$ with no space cost.
    \item In temporal learning module (\equref{equ:TDL}), we use MLP backbones, which lead to $\mathcal{O}\left(MNT_{l-1}T_{l}\right)$ time and $\mathcal{O}\left(MT_{l-1}T_{l}\right)$ space complexity in layer $l$.
    \item In channel learning module (\equref{equ:CCL}), the space and time complexities are trivial since we regard the hidden dimension $D$ as a constant.
    \item Other components include the embedding layer ($\mathcal{O}\left(MNTC\right)$ for time and $\mathcal{O}\left(MC\right)$ for space) and the normalization layer ($\mathcal{O}\left(MNT_{l}\right)$ for time and $\mathcal{O}\left(MT_{l}\right)$ for space in layer $l$).
\end{itemize}

In summary, the time spent on all components is $\mathcal{O}\left(MNTC+M^2N^2+\sum_{l=1}^L(M^2N^2T_{l}+MNT_{l-1}T_{l})\right)$, and the space cost is $\mathcal{O}\left(MC+M^2N+M\sum_{l=1}^L(T_{l-1}T_{l}+T_{l})\right)$.

We further reduce the complexity by letting $T_{l}=\frac{1}{2}T_{l-1}$ and assuming $L<log(T)$, which leads to the overall time complexity $\mathcal{O}\left(MNT(MN+T+C)\right)$ and the overall space complexity $\mathcal{O}\left(M(MN+T^2+C)\right)$.

Compared with most previous methods, SimMST theoretically has a much lower time and space complexity and can empirically outperform the state-of-the-arts with fewer parameters in the experiments.

\section{Experiments}

In this section, we conduct extensive experiments on three real-world multi-mode datasets to evaluate the model performance and present a detailed analysis.

\begin{table}[!htbp]
    \centering
    \normalsize
    \caption{Statistics of the datasets. Density refers to the proportion of nonzero entries of each data.}
    \label{tab:DatasetStatistics}
    \resizebox{\columnwidth}{!}{
        \setlength{\tabcolsep}{1.5mm}{
            \begin{tabular}{cc|cc|cc|c}
            \hline
            City & Mode & Start Time & End Time & \makecell[c]{Number of\\regions} & \makecell[c]{Number of\\Records} & Density \\
            \hline
            \multirow{2}{*}{NYC} & bike & \multirow{2}{*}{4/1/2016} & \multirow{2}{*}{6/30/2016} & \multirow{2}{*}{266} & 7,371,175 & 0.2045 \\
            & taxi &  &  &  & 61,467,659 & 0.9458 \\
            \hline
            \multirow{2}{*}{Chicago} & bike & \multirow{2}{*}{4/1/2016} & \multirow{2}{*}{6/30/2016} & \multirow{2}{*}{247} & 2,140,950 & 0.2739 \\
            & taxi &  &  &  & 9458490 & 0.1803 \\
            \hline
            \multirow{2}{*}{Beijing} & railway & \multirow{2}{*}{7/1/2017} & \multirow{2}{*}{9/30/2017}  & \multirow{2}{*}{123} & 805,136,046 & 0.7908 \\
            & bus &  &  &  & 416,870,574 & 0.8735 \\
            \hline
            \end{tabular}
        }
    }
\end{table}

\begin{table*}[!t]
    \centering
    \normalsize
    \caption{Overall prediction performance of all the methods. The best and the second-best performance are boldfaced and underlined. Lower MAE, lower RMSE, and higher CORR metrics correspond to better performance.}
    \label{tab:performance}
    \resizebox{1.6\columnwidth}{!}
    {
            \begin{tabular}{c|c|ccc|ccc|ccc}
            \hline
            \multirow{2}{*}{Dataset} & \multirow{2}{*}{Method} & \multicolumn{3}{c|}{$H=3$} & \multicolumn{3}{c|}{$H=6$} & \multicolumn{3}{c}{$H=12$} \\
            \cline{3-11}
            & & MAE$\downarrow$ & RMSE$\downarrow$ & CORR$\uparrow$ & MAE$\downarrow$ & RMSE$\downarrow$ & CORR$\uparrow$ & MAE$\downarrow$ & RMSE$\downarrow$ & CORR$\uparrow$ \\
            \hline
            \multirow{9}{*}{\makecell[c]{NYC-Bike}}
            & STGCN & 1.172 & 4.048 & 0.209 & 1.252 & 4.347 & 0.205 & 1.434 & 5.162 & 0.195 \\
            & GWNet & 1.131 & 4.374 & 0.208 & 1.294 & 4.890 & 0.203 & 1.471 & 5.609 & 0.189 \\
            & MTGNN & 0.960 & \underline{3.158} & 0.223 & 1.062 & 3.543 & 0.218 & 1.199 & 4.123 & 0.208 \\
            & ESG & 0.994 & 3.246 & 0.220 & 1.047 & \underline{3.442} & 0.217 & 1.206 & 4.195 & 0.206 \\
            & FC-GAGA & 1.021 & 3.375 & 0.222 & 1.135 & 3.887 & 0.216 & 1.177 & 3.972 & 0.211 \\
            & MTS-Mixer & 1.094 & 3.721 & 0.216 & 1.219 & 4.335 & 0.209 & 1.336 & 4.782 & 0.200 \\
            \cline{2-11}
            & MOHER & 1.034 & 3.488 & 0.213 & 1.110 & 3.801 & 0.207 & 1.213 & 4.197 & 0.200 \\
            & CoGNN & \underline{0.953} & 3.188 & \underline{0.224} & \underline{1.025} & 3.454 & \underline{0.221} & \underline{1.117} & \underline{3.828} & \underline{0.215} \\
            & SimMST & \textbf{0.915} & \textbf{3.030} & \textbf{0.226} & \textbf{0.965} & \textbf{3.210} & \textbf{0.224} & \textbf{1.051} & \textbf{3.576} & \textbf{0.218} \\
            \hline
            \multirow{8}{*}{\makecell[c]{NYC-Taxi}}
            & STGCN & 6.072 & 11.089 & 0.804 & 6.273 & 11.668 & 0.796 & 6.966 & 13.482 & 0.752 \\
            & GWNet & 5.253 & 9.236 & 0.852 & 5.298 & 9.423 & 0.849 & 5.690 & 10.240 & 0.826 \\
            & MTGNN & 4.772 & 8.320 & 0.872 & 4.956 & 8.850 & 0.863 & 5.368 & 9.723 & 0.849 \\
            & ESG & 5.135 & 8.963 & 0.859 & 5.254 & 9.347 & 0.852 & 5.692 & 10.112 & 0.834 \\
            & FC-GAGA & 4.961 & 8.726 & 0.862 & 5.163 & 9.313 & 0.853 & 5.552 & 9.930 & 0.840 \\
            & MTS-Mixer & 5.267 & 9.302 & 0.851 & 5.311 & 9.433 & 0.850 & 5.655 & 10.227 & 0.839 \\
            \cline{2-11}
            & MOHER & 5.317 & 9.211 & 0.829 & 5.725 & 10.038 & 0.802 & 6.397 & 11.314 & 0.761 \\
            & CoGNN & \underline{4.681} & \underline{8.125} & \underline{0.873} & \underline{4.898} & \underline{8.728} & \underline{0.865} & \underline{5.292} & \underline{9.641} & \underline{0.851} \\  
            & SimMST & \textbf{4.613} & \textbf{7.984} & \textbf{0.874} & \textbf{4.802} & \textbf{8.498} & \textbf{0.868} & \textbf{5.089} & \textbf{9.161} & \textbf{0.857} \\
            \hline
            \multirow{8}{*}{\makecell[c]{Chicago-Bike}}
            & STGCN & 0.803 & 2.566 & 0.308 & 0.849 & 2.771 & 0.275 & 0.907 & 3.052 & 0.269 \\
            & GWNet & 1.428 & 5.175 & 0.312 & 1.431 & 5.177 & 0.236 & 1.420 & 5.132 & 0.228 \\
            & MTGNN & \underline{0.710} & \textbf{1.842} & \underline{0.348} & \underline{0.744} & \underline{2.005} & \underline{0.347} & 0.802 & 2.343 & \underline{0.326} \\
            & ESG & 0.741 & 2.053 & 0.337 & 0.774 & 2.166 & 0.317 & 0.870 & 2.750 & 0.287 \\
            & FC-GAGA & 0.775 & 2.064 & 0.333 & 0.829 & 2.425 & 0.308 & 0.883 & 2.679 & 0.258 \\
            & MTS-Mixer & 0.820 & 2.494 & 0.345 & 0.848 & 2.644 & 0.330 & 0.912 & 2.867 & 0.293 \\
            \cline{2-11}
            & MOHER & 0.790 & 2.348 & 0.320 & 0.835 & 2.576 & 0.291 & 0.883 & 2.869 & 0.286 \\
            & CoGNN & 0.724 & 1.996 & 0.342 & 0.752 & 2.114 & 0.326 & \underline{0.800} & \underline{2.323} & 0.309 \\
            & SimMST & \textbf{0.708} & \underline{1.876} & \textbf{0.368} & \textbf{0.736} & \textbf{2.004} & \textbf{0.360} & \textbf{0.776} & \textbf{2.207} & \textbf{0.351} \\
            \hline
            \multirow{8}{*}{\makecell[c]{Chicago-Taxi}}
            & STGCN & 1.302 & 4.904 & 0.312 & 1.403 & 5.292 & 0.253 & 1.522 & 5.735 & 0.286 \\
            & GWNet & 1.260 & 4.807 & 0.258 & 1.369 & 5.346 & 0.245 & 1.514 & 6.014 & 0.245 \\
            & MTGNN & 1.057 & 3.759 & 0.307 & 1.131 & \underline{4.071} & 0.290 & 1.222 & 4.512 & 0.285 \\
            & ESG & 1.107 & 3.929 & 0.319 & 1.165 & 4.182 & 0.296 & 1.349 & 5.134 & 0.261 \\
            & FC-GAGA & 1.137 & 4.117 & 0.313 & 1.275 & 4.973 & 0.280 & 1.432 & 5.548 & 0.259 \\
            & MTS-Mixer & 1.232 & 4.588 & 0.302 & 1.305 & 4.950 & 0.298 & 1.462 & 5.718 & 0.280 \\
            \cline{2-11}
            & MOHER & 1.127 & 3.934 & 0.303 & 1.241 & 4.459 & 0.278 & 1.391 & 5.101 & 0.268 \\
            & CoGNN & \underline{1.025} & \underline{3.596} & \underline{0.332} & \underline{1.107} & 4.074 & \underline{0.333} & \underline{1.189} & \underline{4.453} & \underline{0.310} \\
            & SimMST & \textbf{1.001} & \textbf{3.481} & \textbf{0.342} & \textbf{1.076} & \textbf{3.921} & \textbf{0.335} & \textbf{1.159} & \textbf{4.339} & \textbf{0.328} \\
            \hline
            \multirow{8}{*}{\makecell[c]{Beijing-Railway}}
            & STGCN & 114.684 & 270.681 & 0.952 & 146.426 & 439.151 & 0.915 & 124.038 & 284.024 & 0.951 \\
            & GWNet & 81.743 & 199.084 & 0.971 & 90.111 & 245.067 & 0.962 & \underline{96.837} & \underline{255.000} & \underline{0.962} \\
            & MTGNN & \underline{71.674} & \textbf{166.712} & \textbf{0.977} & \underline{82.079} & \textbf{218.255} & \underline{0.968} & 113.155 & 419.650 & 0.933 \\
            & ESG & 78.350 & 204.110 & 0.966 & 90.826 & 245.604 & 0.950 & 112.458 & 333.874 & 0.941 \\
            & FC-GAGA & 90.356 & 273.169 & 0.964 & 120.048 & 544.727 & 0.916 & 110.479 & 378.218 & 0.951 \\
            & MTS-Mixer & 95.116 & 252.849 & 0.966 & 116.973 & 398.508 & 0.937 & 126.121 & 384.055 & 0.939 \\
            \cline{2-11}
            & MOHER & 139.898 & 430.064 & 0.909 & 165.280 & 549.150 & 0.889 & 150.780 & 408.103 & 0.920 \\
            & CoGNN & 74.776 & 180.763 & 0.975 & 88.649 & 251.715 & 0.962 & 102.099 & 347.678 & 0.948 \\
            & SimMST & \textbf{65.039} & \underline{170.731} & \underline{0.976} & \textbf{75.331} & \underline{231.130} & \textbf{0.970} & \textbf{77.567} & \textbf{234.012} & \textbf{0.969} \\
            \hline
            \multirow{8}{*}{\makecell[c]{Beijing-Bus}}
            & STGCN & 52.971 & 93.294 & 0.956 & 58.835 & 113.180 & 0.943 & 62.140 & 117.252 & 0.942 \\
            & GWNet & 48.674 & 83.301 & 0.965 & 55.458 & 108.946 & 0.951 & 56.155 & 107.763 & 0.951 \\
            & MTGNN & \underline{41.420} & \underline{72.097} & \underline{0.974} & \underline{44.884} & \underline{84.857} & \underline{0.968} & \underline{48.328} & 94.591 & \underline{0.963} \\
            & ESG & 46.265 & 80.764 & 0.958 & 51.428 & 97.420 & 0.941 & 52.016 & \underline{90.205} & 0.945 \\
            & FC-GAGA & 44.488 & 76.874 & 0.969 & 51.740 & 105.501 & 0.955 & 55.735 & 114.944 & 0.947 \\
            & MTS-Mixer & 49.211 & 91.957 & 0.958 & 54.258 & 110.886 & 0.949 & 56.259 & 115.168 & 0.946 \\
            \cline{2-11}
            & MOHER & 56.589 & 109.901 & 0.940 & 67.910 & 144.560 & 0.915 & 66.926 & 123.500 & 0.927 \\
            & CoGNN & 42.136 & 73.651 & 0.972 & 46.772 & 88.065 & 0.966 & 51.011 & 99.085 & 0.960 \\
            & SimMST & \textbf{38.065} & \textbf{67.629} & \textbf{0.975} & \textbf{41.488} & \textbf{75.248} & \textbf{0.972} & \textbf{44.121} & \textbf{78.761} & \textbf{0.970} \\
            \hline
            \end{tabular}
        }
\end{table*}

\subsection{Descriptions of Datasets}
We use three multi-mode traffic datasets for experiments, which record the traffic demands of three cities (New York City, Chicago, and Beijing) with multiple modes (bikes, taxis, buses, and railways):
\begin{itemize}
    \item \textbf{NYC-Bike \& NYC-Taxi}. We use the Bike Sharing System to obtain the demands of bikes in New York City, which collects sharing bike orders and shares them online\footnote{\url{https://ride.citibikenyc.com/system-data}}. Each order record contains information on pick-up/drop-off time, latitude, and longitude. We use the dataset that contains taxi trip records to get the taxi demands\footnote{\url{https://data.cityofnewyork.us/Transportation/2016-Yellow-Taxi-Trip-Data/k67s-dv2t}}. We select records from April 2016 to June 2016, and obtain 7,371,175 bike records and 61,467,659 taxi records in total.
    
    \item \textbf{Chicago-Bike \& Chicago-Taxi}. We obtain the records of bikes and taxis from Divvy\footnote{\url{https://ride.divvybikes.com/how-it-works}}, a Chicago's bike share system, and The City of Chicago\footnote{\url{https://data.cityofchicago.org/Transportation/Taxi-Trips/wrvz-psew}}. Each trip record is anonymized and includes trip start/end time and start/end location. In total, we select 2,140,950 bike records and 9,458,490 taxi records from April 2016 to June 2016 to do experiments.
    
    \item \textbf{Beijing-Railway \& Beijing-Bus}. We use the Beijing Identity Card Records dataset, which contains public transport records in Beijing of railways and buses, to obtain the demands of railways and buses. We finally choose 805,136,046 railway trip records and 416,870,574 bus trip records from July 2017 to September 2017 for experiments.
\end{itemize}
We partition cities into regions and time into intervals. For each region and time interval, we filter trips with pick-up and drop-off locations and timestamps within that region and interval. We use the number of these trips as the pick-up demands and drop-off demands, respectively. This process is performed separately for each mode and the data is fed into the corresponding mode-specific modules. Statistics of the datasets are shown in \tabref{tab:DatasetStatistics}. Following \cite{Wu_MTGNN_Connecting_2020}, we split each dataset into the training set ($70\%$), validation set ($15\%$), and test set ($15\%$) in chronological order. Both the historical and future sequence lengths are set to 12 and the time slot interval is 30 minutes. The task is to predict future traffic demands for each mode given historical multi-mode traffic demands.

\subsection{Baselines}

We compare our method with the following baselines, including single-mode spatial-temporal data modeling methods (\textbf{STGCN}, \textbf{GWNet}, \textbf{MTGNN}, \textbf{ESG}, \textbf{FC-GAGA} and \textbf{MTS-Mixer}) and methods for multi-mode spatial-temporal data modeling (\textbf{MOHER} and \textbf{CoGNN}):

\begin{itemize}
    \item \textbf{STGCN} \cite{Yu_STGCN_SpatioTemporal_2018} uses pure convolutions to model spatial-temporal data on predefined graphs.
    \item \textbf{GWNet} \cite{Wu_GraphWaveNet_Graph_2019} introduces an adaptively learned adjacency matrix, along with predefined graphs, to capture the spatial relationships between nodes.
    \item \textbf{MTGNN} \cite{Wu_MTGNN_Connecting_2020} is a widely-adopted baseline for spatial-temporal data modeling, which effectively learns inherent spatial relationships.
    \item \textbf{ESG} \cite{Ye_ESG_Learning_2022} constructs multiple evolutionary graphs to model the dynamic interactions of time series.
    \item \textbf{FC-GAGA} \cite{Oreshkin_FC-GAGA_FCGAGA_2021} extends \cite{Oreshkin_N-BEATS_NBEATS_2020} with learnable time gates and graph gates for spatial-temporal modeling with fully connected layers.
    \item \textbf{MTS-Mixer} \cite{Li_MTS-Mixers_MTSMixers_2023} uses factorized multi-layer perceptrons to effectively capture the temporal and channel interactions.
    \item \textbf{MOHER} \cite{Zhou_MOHER_Modeling_2021} explores the correlations and differences between multiple modes by using the predefined geographical proximity graphs and the POI similarity graphs.
    \item \textbf{CoGNN} \cite{Liu_CoGNN_CoPrediction_2021} constructs a heterogeneous graph to learn the spatial relationships of multiple transportation modes.
\end{itemize}

Since single-mode method is not specifically designed for multi-mode datasets, we train an individual model for each mode, e.g. training two individual STGCN instances respectively on NYC-Bike dataset and NYC-Taxi dataset.

\subsection{Experimental Setup}
We keep the basic settings of baselines and our method to be identical (e.g., learning rate as 0.001, hidden dimension as 32). We use grid search to find the optimal settings for model-specific hyper-parameters to get more reliable results, such as model layers and dropout rate.
We train all the methods for 1000 epochs with batch size 128 and early stop patience 100. 
For our method, top 20 entries of each row in matrix $\Tilde{\bm{A}}$ are selected and the dimension $D_{emb}$ of region embedding is 40.
We stack 3 layers to build our model.
The hidden dimension $D$ is set to 32 for all modules as well as the time semantic embedding.
Following \cite{Ye_ESG_Learning_2022}, we use MAE, RMSE, and CORR as the evaluation metrics. All the experiments are conducted on a Centos machine equipped with Intel(R) Xeon(R) CPU E5-2667 and NVIDIA TITAN Xp with 12GB memory, with python 3.10 and PyTorch 1.12 \cite{Paszke_PyTorch_PyTorch_2019}. 

\subsection{Overall Performance}
\label{section:overall}

\begin{figure}[!htbp]
    \centering
    \includegraphics[width = \columnwidth]{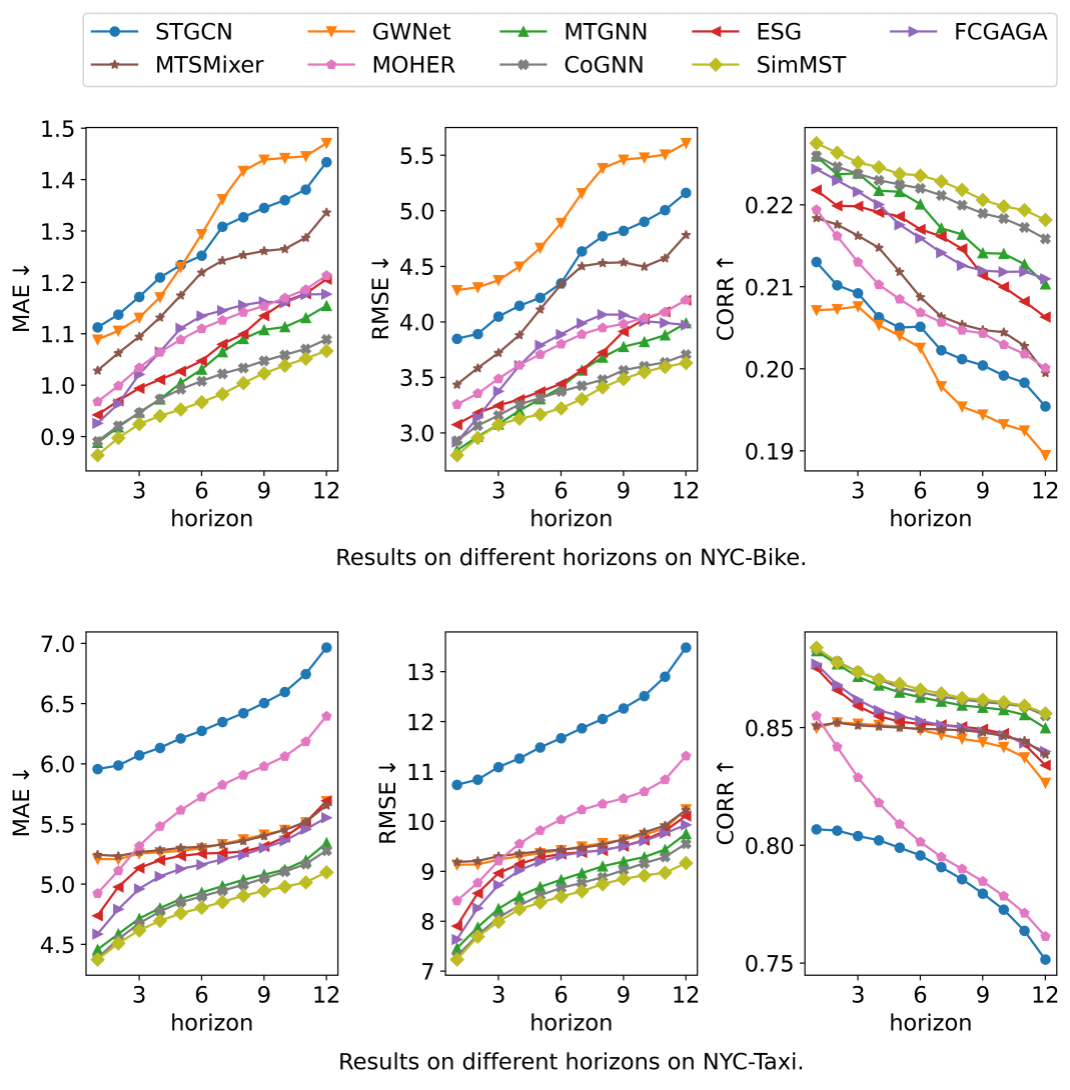}
    \caption{Prediction results on different horizons on NYC dataset.}
    \label{fig:ExpHorizon}
\end{figure}

We show the overall prediction performance of all the methods in \tabref{tab:performance} and present the results on different horizons in \figref{fig:ExpHorizon}. We observe that our approach achieves the best performance in most cases even if its framework is quite simple, and several conclusions can be further summarized:
Firstly, \textit{adaptively learning spatial relationships is essential for spatial-temporal data modeling}. Compared with STGCN, GWNet, and MTGNN achieve better performance, indicating the predefined relation matrix can not fully capture the spatial relationships between regions, while the learned matrix reflects the inherent relationship more effectively.
Secondly, \textit{modeling the relationships between multiple modes is beneficial}. Although both STGCN and MOHER use predefined graphs, MOHER surpasses STGCN in most cases by considering multiple modes. Similarly, though CoGNN and MTGNN automatically learn the spatial relationships, CoGNN often performs better than MTGNN as it constructs a heterogeneous graph to capture the impacts from different modes.
Finally, \textit{SimMST consistently outperforms the existing methods in most cases by using a quite simple architecture}, which demonstrates its effectiveness in learning the temporal dependencies, cross-mode spatial relationships, and channel correlations in multi-mode spatial-temporal data.

\subsection{Importance of Modeling Multi-mode Relationships}
\label{section:Multi-mode}

In this part, we aim to verify 1) the importance of modeling the relationships of multiple modes; and 2) the generalizability and efficacy of the proposed cross-mode spatial relationships learning (CSRL) component.

\begin{figure}[h]
    \centering
    \includegraphics[width = \columnwidth]{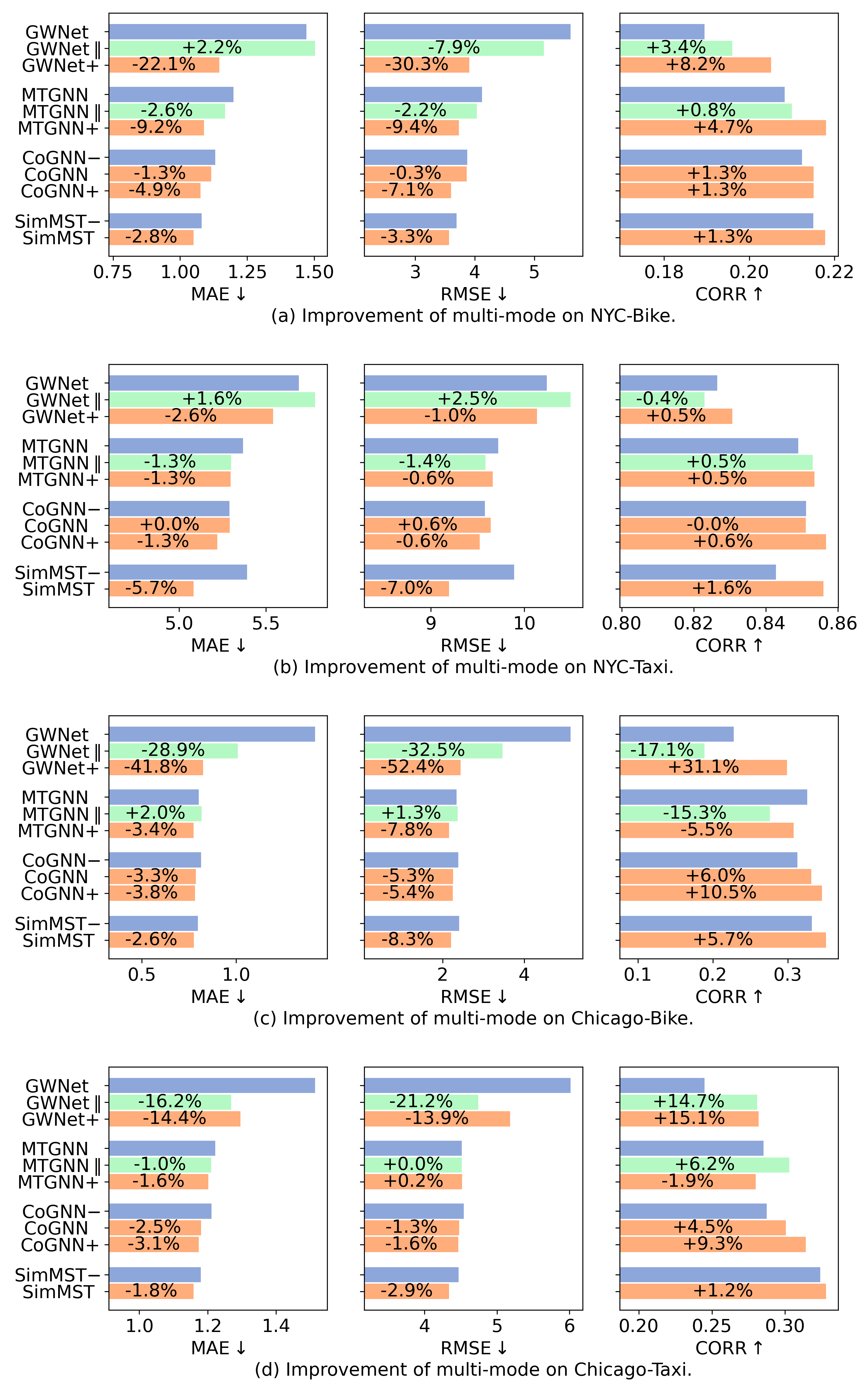}
    \caption{Performance ($H=12$) of different variants on the NYC and Chicago datasets. Blue/Mint/Orange bars denote the performance of methods without modeling multi-mode relationships / with naive combinations of multiple modes / with the capturing of multi-mode relationships. Metric differences compared to blue bars are marked. CoGNN+ refers to substituting its original multi-mode modeling module with our proposed CSRL component.}
    \label{fig:ExpMultimode}
\end{figure}

We first choose GWNet and MTGNN as two representative single-mode methods, which achieve relatively better performance. Then, we provide two ways to adapt them for handling multi-mode spatial-temporal data. One naive way is to treat the observations from multiple modes as different features and \textit{combine} them together as the inputs for each model. We denote such variants as GWNet $\|$ and MTGNN $\|$. Another way is to \textit{integrate} our CSRL component into the models and enable them to capture multi-mode relationships. We denote these variants as GWNet+ and MTGNN+.
Also, for multi-mode spatial-temporal data modeling methods, we \textit{remove} the multi-mode modeling module in CoGNN and the proposed SimMST, and denote the variants CoGNN- and SimMST-.
Further, we \textit{substitute} the multi-mode modeling module in CoGNN with the proposed CSRL component and denote it as CoGNN+.
We show the performance of all the above methods on the NYC and Chicago datasets in \figref{fig:ExpMultimode}.


From \figref{fig:ExpMultimode}, we find that 
1) combining the observations from multiple modes as input features can help the single-mode methods perform better in most cases, which shows the importance of considering cross-mode information;
2) the proposed cross-mode spatial relationships learning (CSRL) component can effectively improve the model performance further and leads to better performance than the naive combinations of multiple modes, which indicates the necessity of designing customized modules and demonstrates its effectiveness and generalizability;
3) after removing the multi-mode relationship module in CoGNN and our SimMST (CoGNN- v.s. CoGNN and SimMST- v.s. SimMST), the results become worse, which shows the necessity of learning the relationships of multiple modes;
4) by substituting the original multi-mode modeling module of CoGNN with our CSRL component (CoGNN v.s. CoGNN+), CoGNN achieves higher performance, thus demonstrating the efficacy of our proposed CSRL component.


\begin{figure*}[h]
    \centering
    \includegraphics[width = 1.55\columnwidth]{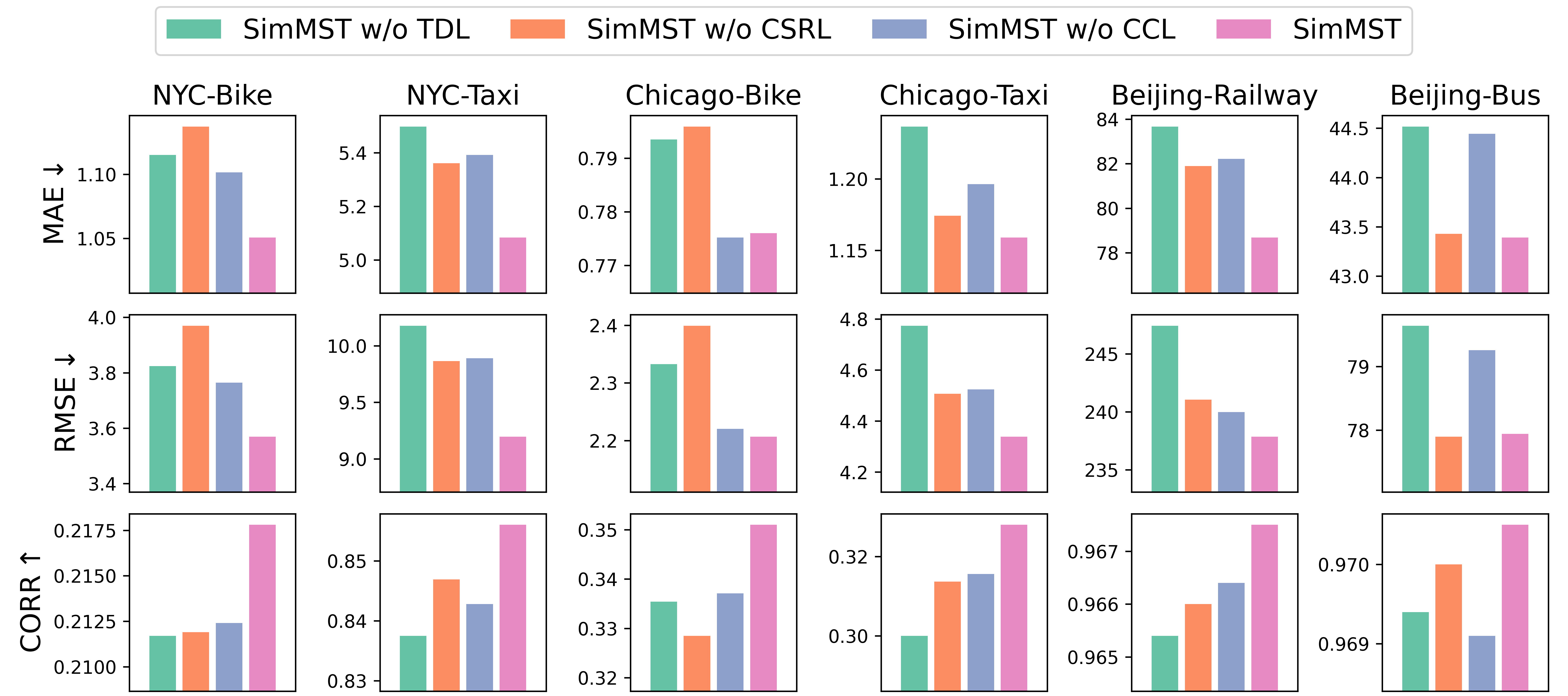}
    \caption{Effects of different components on all three datasets. Performances with $H=12$ are reported.}
    \label{fig:Ablation}
\end{figure*}

\subsection{Ablation Study}

We further conduct an ablation study to validate the effectiveness of the proposed Temporal Dependencies Learning (TDL) component, Cross-mode Spatial Relationships Learning (CSRL) component, and Channel Correlations Learning (CCL) component. We remove TDL, CSRL, and CCL components from the proposed SimMST, and denote the remaining parts as SimMST w/o TDL, SimMST w/o CSRL, and SimMST w/o CCL, respectively. The results of different variants on all three datasets are shown in \figref{fig:Ablation}.

From \figref{fig:Ablation}, we observe that SimMST achieves the best performance when using all three components. Removing any component will degrade the performance. To be specific, TDL captures the evolutionary pattern of each region in the temporal dimension. CSRL propagates the information between regions under multiple modes to simultaneously exploit the cross-mode and spatial relationships. CCL enhances the model capacity by learning channel correlations.
Hence, the contribution of each component is verified.

\subsection{Model Efficiency Comparison}
Compared with previous methods, another advantage of our approach stems from its simple framework with lower model complexity. Therefore, in this part, we validate the efficiency of our approach by comparing the parameter size and the inference time with baselines. Moreover, we conduct experiments on some variants of SimMST by using seasonality configuration (denoted as \textit{SimMST-Season}), recurrent-based module (e.g. GRU, denoted as \textit{SimMST-RNN}) and Attention-based module (e.g. Transformer, denoted as \textit{SimMST-Attn}) as TDL implementations.
The experimental results on all the datasets are shown in \tabref{tab:efficiency}.

\begin{table}[!htbp]
    \centering
    \normalsize
    \caption{Comparison of parameter size, inference time, and performance on all datasets.}
    \label{tab:efficiency}
    \resizebox{\columnwidth}{!}{
        \setlength{\tabcolsep}{1.0mm}{
            \begin{tabular}{c|c|ccc|cc|cc}
            \hline
            Dataset & Methods & \makecell[c]{Bike/Railway\\MAE-12} & \makecell[c]{Taxi/Bus\\MAE-12} & \makecell[c]{Relative\\Performance} & \makecell[c]{Parameter\\Size} & \makecell[c]{Relative\\Size} & \makecell[c]{Inference\\Time\\(s/epoch)}& \makecell[c]{Relative\\Time} \\
            \hline
            \multirow{8}{*}{NYC}
            & MOHER & 1.213 & 6.397 & -16.9\% & 92,492 & -58.0\% & 1.77 & +14.4\% \\
            & CoGNN & 1.117 & 5.292 & -4.9\% & 932,632 & +323.7\% & 2.26 & +46.3\% \\
            & GWNet+ & 1.168 & 5.545 & -9.2\% & 538,332 & +144.5\% & 6.74 & +335.9\% \\
            & MTGNN+ & 1.089 & 5.297 & -3.7\% & 995,876 & +352.4\% & 2.19 & +41.8\% \\
            \cline{2-9}
            & SimMST-Season & 1.071 & 5.100 & -1.1\% & 226,782 & +3.0\% & 1.51 & -2.5\% \\
            & SimMST-RNN & 1.064 & 5.122 & -0.9\% & 295,694 & +34.3\% & 1.95 & +26.4\% \\
            & SimMST-Attn & 1.105 & 5.099 & -2.5\% & 424,718 & +92.9\% & 4.08 & +163.7\% \\
            \cline{2-9}
            & SimMST & 1.051 & 5.089 & - & 220,140 & - & 1.55 & - \\
            \hline
            \multirow{8}{*}{Chicago}
            & MOHER & 0.883 & 1.391 & -14.4\% & 92,492 & -57.5\% & 1.67 & +17.3\% \\
            & CoGNN & 0.800 & 1.189 & -2.7\% & 874,264 & +302.1\% & 2.07 & +45.4\% \\
            & GWNet+ & 0.830 & 1.314 & -9.1\% & 526,892 & +142.4\% & 6.49 & +355.7\% \\
            & MTGNN+ & 0.773 & 1.202 & -1.6\% & 941,004 & +332.8\% & 2.16 & +51.9\% \\
            \cline{2-9}
            & SimMST-Season & 0.779 & 1.147 & +0.3\% & 224,046 & +3.1\% & 1.40 & -1.5\% \\
            & SimMST-RNN & 0.798 & 1.170 & -1.8\% & 409,134 & +88.2\% & 1.78 & +25.3\% \\
            & SimMST-Attn & 0.782 & 1.157 & -0.2\% & 425,262 & +95.6\% & 3.76 & +164.1\% \\
            \cline{2-9}
            & SimMST & 0.776 & 1.159 & - & 217,404 & - & 1.42 & - \\
            \hline
            \multirow{8}{*}{Beijing}
            & MOHER & 150.780 & 66.926 & -41.3\% & 92,492 & -53.6\% & 0.80 & +0.4\% \\
            & CoGNN & 102.099 & 51.011 & -18.8\% & 493,336 & +147.2\% & 1.08 & +35.7\% \\
            & GWNet+ & 103.283 & 50.351 & -18.6\% & 536,812 & +169.0\% & 5.21 & +557.3\% \\
            & MTGNN+ & 108.696 & 50.291 & -20.5\% & 582,892 & +192.1\% & 1.24 & +56.7\% \\
            \cline{2-9}
            & SimMST-Season & 77.939 & 42.333 & +1.9\% & 206,190 & +3.3\% & 0.80 & +0.7\% \\
            & SimMST-RNN & 77.470 & 42.196 & +2.3\% & 391,278 & +96.1\% & 0.95 & +20.0\% \\
            & SimMST-Attn & 88.881 & 46.068 & -8.5\% & 407,406 & +104.2\% & 1.96 & +147.4\% \\
            \cline{2-9}
            & SimMST & 77.567 & 44.121 & - & 199,548 & - & 0.79 & - \\
            \hline
            \end{tabular}
        }
    }
\end{table}

Concretely, on the NYC/Chicago/Beijing dataset, the proposed SimMST achieves 109.6\%/ 117.6\%/ 162.5\% accelerations in time, 65.6\%/ 64.3\%/ 53.2\% reductions in space, and 8.4\%/ 5.5\%/ 33.3\% improvements on average.
The significant decreases in inference time and parameter size demonstrate the efficiency and lightweight property of our approach. Though our SimMST additionally learns the multi-mode relation matrices, it brings better performance and the overall simple framework still guarantees that SimMST enjoys a faster speed.

Moreover, compared with complex components such as RNN and Attention, SimMST with MLPs achieves comparable performance despite its simple structure. This suggests that multi-mode spatial-temporal data can be well modeled by a simple framework and complex modules may be unnecessary.


\subsection{Visualization}

\begin{figure}[h]
    \centering
    \includegraphics[width = \columnwidth]{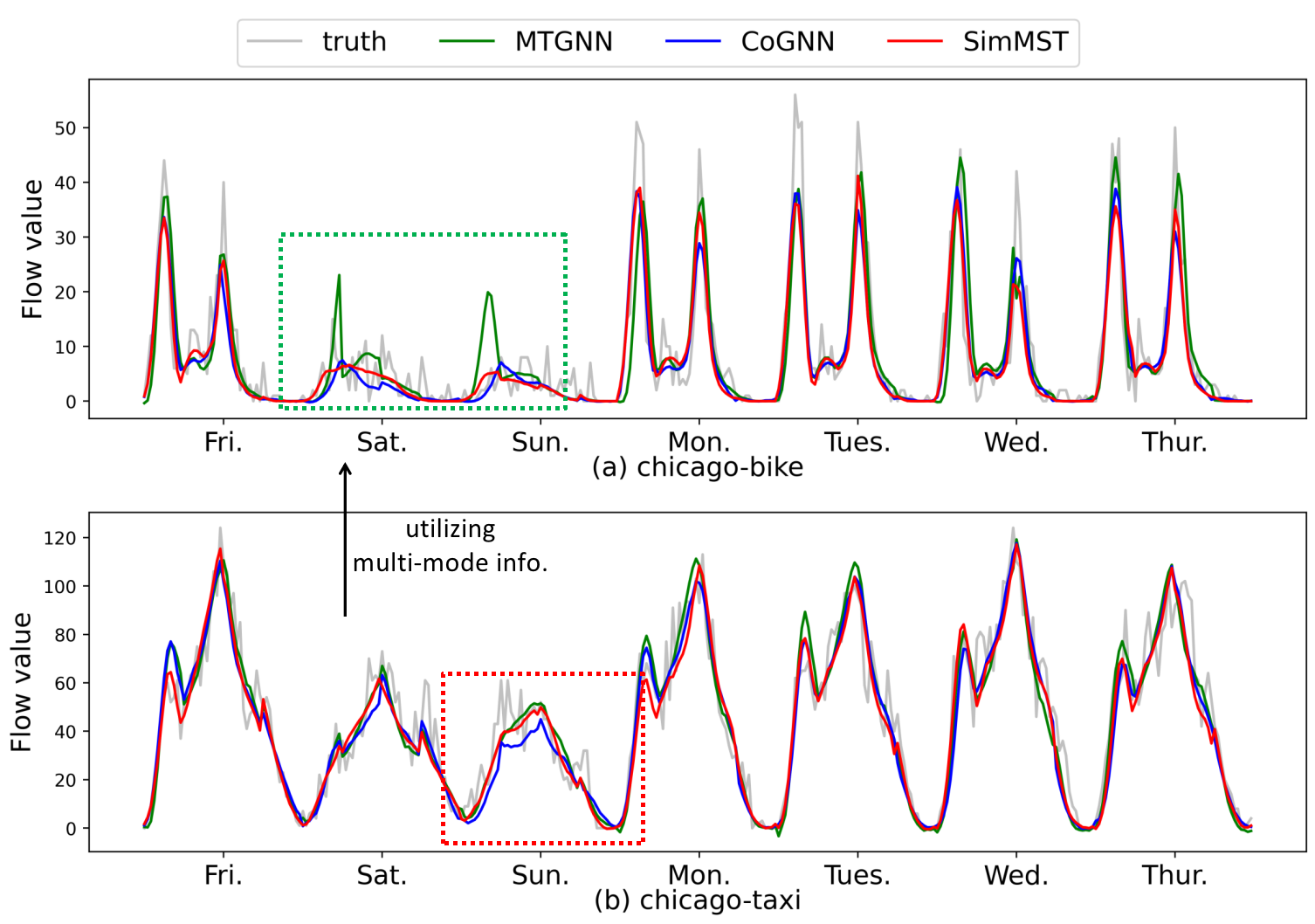}
    \caption{Forecasting visualizations of a case region (id \#242) on Chicago dataset. Grey curves refer to the truth observations and green, blue, and red curves respectively refer to the predictions of MTGNN, CoGNN, and SimMST. Only the first 7 days in the test set (6/17/2016 - 6/23/2016) are shown.}
    \label{fig:Visualization}
\end{figure}

We provide a case study to demonstrate the importance of modeling multi-mode relationships and the effectiveness of the proposed method. \figref{fig:Visualization} shows the future traffic demands predicted by MTGNN, CoGNN, and SimMST and the ground truths of a case region on the Chicago dataset.


Despite all methods largely approximate the truth demands, their prediction details are not identical:
1) Compared with methods with multi-mode modeling, single-mode method fails to correctly predict the bike demands on weekends. Specifically, MTGNN mistakenly predicts a peak in the morning on weekends (see green box in the figure), while multi-mode methods (CoGNN and SimMST) can utilize the information from taxi demands and provide more accurate predictions.
2) SimMST is generally more accurate and robust than CoGNN, which can be inferred from the green and red boxes in the figure. With fewer parameters, SimMST can better avoid overfitting the noises of training data and yield smoother predictions.




\section{Related Work}

In this section, we review the related work and show the differences between previous studies and our work, including multi-mode spatial-temporal data modeling methods, and simple frameworks in different fields.

\subsection{Multi-mode Spatial-Temporal Data Modeling}
Due to the importance of spatial-temporal data, a number of efforts have been made on modeling spatial-temporal data. Most of the existing methods mainly focus on learning from a single mode  \cite{Li_DCRNN_Diffusion_2018,Yu_STGCN_SpatioTemporal_2018,DBLP:conf/kdd/YiZWLZ18,Wu_GraphWaveNet_Graph_2019,Guo_ASTGCN_Attention_2019,wen2019novel,DBLP:conf/aaai/ChenCXCGF20,DBLP:conf/aaai/ZhengFW020,Wu_MTGNN_Connecting_2020,Oreshkin_FC-GAGA_FCGAGA_2021,Ye_ESG_Learning_2022,Li_MTS-Mixers_MTSMixers_2023}.
For example, \cite{Li_DCRNN_Diffusion_2018} combined diffusion graph convolution network on predefined graphs and gated recurrent units for modeling traffic data. STGCN \cite{Yu_STGCN_SpatioTemporal_2018} applied purely convolutional structures to jointly extract spatial-temporal features for traffic speed forecasting. \cite{Wu_GraphWaveNet_Graph_2019,Wu_MTGNN_Connecting_2020,Bai_AGCRN_Adaptive_2020,Choi_STG-NCDE_Graph_2022} designed GNN-based frameworks to exploit inherent spatial relationships. ESG \cite{Ye_ESG_Learning_2022} constructed multiple evolutionary graph structures to model the dynamic interactions of time series. However, these methods ignore the relationships between the data from multiple modes, which are pervasive in many real-world scenarios.

In recent years, very few researchers investigated the modeling of multi-mode spatial-temporal data. One part of the methods restricted the studied objects on regular grids in the space \cite{Liang_GeoMAN_GeoMAN_2018,Huang_MiST_MiST_2019,Ye_CoST-Net_CoPrediction_2019,Deng_CEST_Pulse_2021}, and thus limited the models' usability in scenarios with irregular objects.
Another part of the methods applied GNNs to propagate information along the connections between regions. STC-GNN \cite{Wang_STC-GNN_SpatioTemporalCategorical_2021} learned pairwise similarity between modes and proposed a spatial-temporal encoder-decoder framework. MOHER \cite{Zhou_MOHER_Modeling_2021} built graphs with predefined geographical and functional similarities and proposed a propagation mechanism to explore both multi-mode and spatial relationships. CoGNN \cite{Liu_CoGNN_CoPrediction_2021} constructed a heterogeneous graph to learn the inherent spatial relationships of multiple transportation modes with spatial-temporal convolutional networks.

We conclude that previous methods either neglect the multi-mode relationships or need sophisticated modules with higher model complexity. In this paper, we simultaneously capture the temporal dependencies, cross-mode spatial relationships, and channel correlations for modeling multi-mode spatial-temporal data with lower space and time complexity.

\subsection{Simple Frameworks in Different Fields}
Recently, there is an increasing interest in designing simple frameworks in a variety of fields. 
In computer vision, \cite{Tolstikhin_MLP-Mixer_MLPMixer_2021}, \cite{Touvron_ResMLP_ResMLP_2021} and \cite{Yu_S2-MLP_S2MLP_2022} abandoned both the convolution and attention operations by presenting MLP-based architectures for image classification. \cite{Zhang_MorphMLP_MorphMLP_2022} further presented two MLP-based layers for intra-frame and inter-frame modeling respectively for video representation learning.
In sequential recommendation, \cite{He_LightGCN_LightGCN_2020} simplified GCN and learned user and item embedding by linear propagation on user-item interaction graphs.
\cite{Zhou_FMLP-Rec_Filterenhanced_2022} replaced the multi-head self-attention layers in Transformer with filter-enhanced MLP layers. MLP4Rec \cite{Li_MLP4Rec_MLP4Rec_2022} developed a tri-directional fusion scheme to coherently capture sequential, cross-channel, and cross-feature correlations by MLPs. \cite{Yu_SFCNTSP_Predicting_2023} present a succinct architecture that is solely built on the fully connected networks without nonlinear activation functions for temporal sets modeling.
For traffic forecasting, FCGAGA \cite{Oreshkin_FC-GAGA_FCGAGA_2021} employed learnable time gates to model temporal seasonality and hard graph gates to model spatial and temporal relationships at the same time with fully connected layers. 
MTS-Mixer \cite{Li_MTS-Mixers_MTSMixers_2023} replaced attention modules with factorized multi-layer perceptrons to capture temporal and channel dependencies.
x
The above methods have shown the potential of simple architectures. In this paper, we design a simple framework for modeling multi-mode spatial-temporal data to bring both effectiveness and efficiency together.

\section{Conclusion}
In this paper, we studied the problem of multi-mode spatial-temporal data modeling. Considering the fact that existing methods either failed to capture the multi-mode relationships or were built on complicated modules, we investigated the possibility of designing a simple framework for such a problem. We presented a cross-mode spatial relationships learning component to capture the correlations between multiple modes, which is general and applicable to existing spatial-temporal data modeling methods. We also utilized MLPs to capture the temporal dependencies and channel correlations, which are simple yet effective. Extensive experiments on real-world multi-mode traffic datasets not only showed the effectiveness and efficiency of our approach but also demonstrated the generability of our cross-mode spatial relationships learning module. Our work encourages future research to rethink the key factors for spatial-temporal data modeling and indicates the potential of simple architectures.



\bibliographystyle{IEEEtran}
\bibliography{references}

 
\begin{IEEEbiography}[{\includegraphics[width=1in,height=1.25in,clip,keepaspectratio]{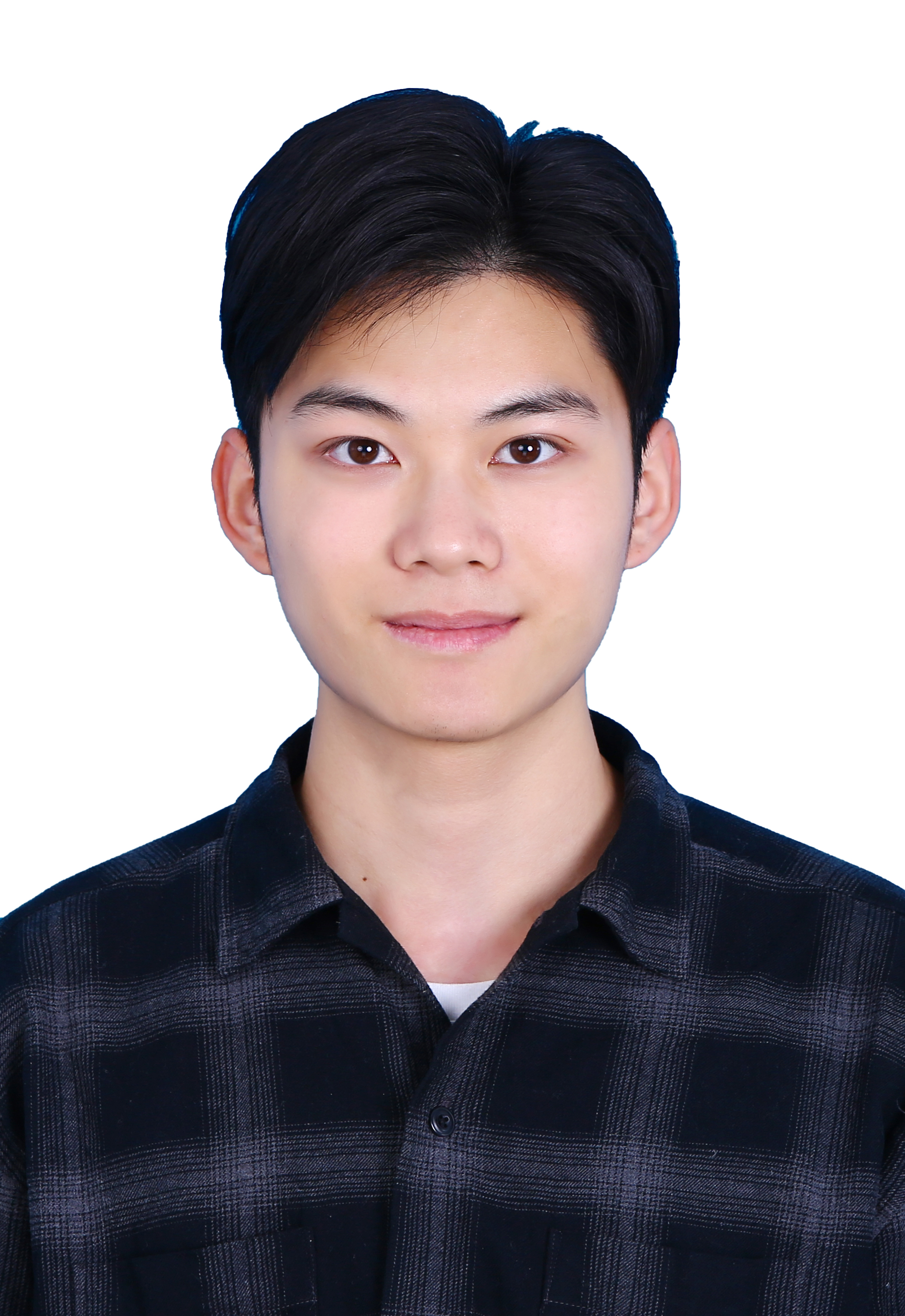}}]
{Zihang Liu} is currently a second-year M.S. student in Computer Science and Engineering from Beihang University. He received the B.S. degree in 2021. His research interests include time-series data mining, recommendation system, and graph neural networks.
\end{IEEEbiography}

\begin{IEEEbiography}[{\includegraphics[width=1in,height=1.25in,clip,keepaspectratio]{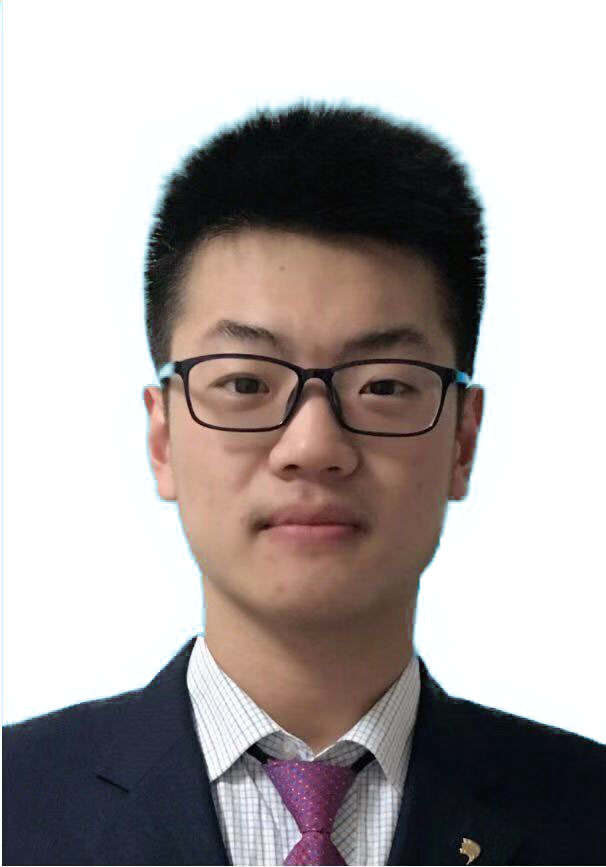}}]
{Le Yu} received the B.S. degree in the School of Computer Science and Engineering at Beihang University, Beijing, China, in 2019. He is currently a fourth-year computer science Ph.D. student in the School of Computer Science and Engineering at Beihang University. His research interests include temporal data mining, machine learning, and graph neural networks.
\end{IEEEbiography}

\begin{IEEEbiography}[{\includegraphics[width=1in,height=1.25in,clip,keepaspectratio]{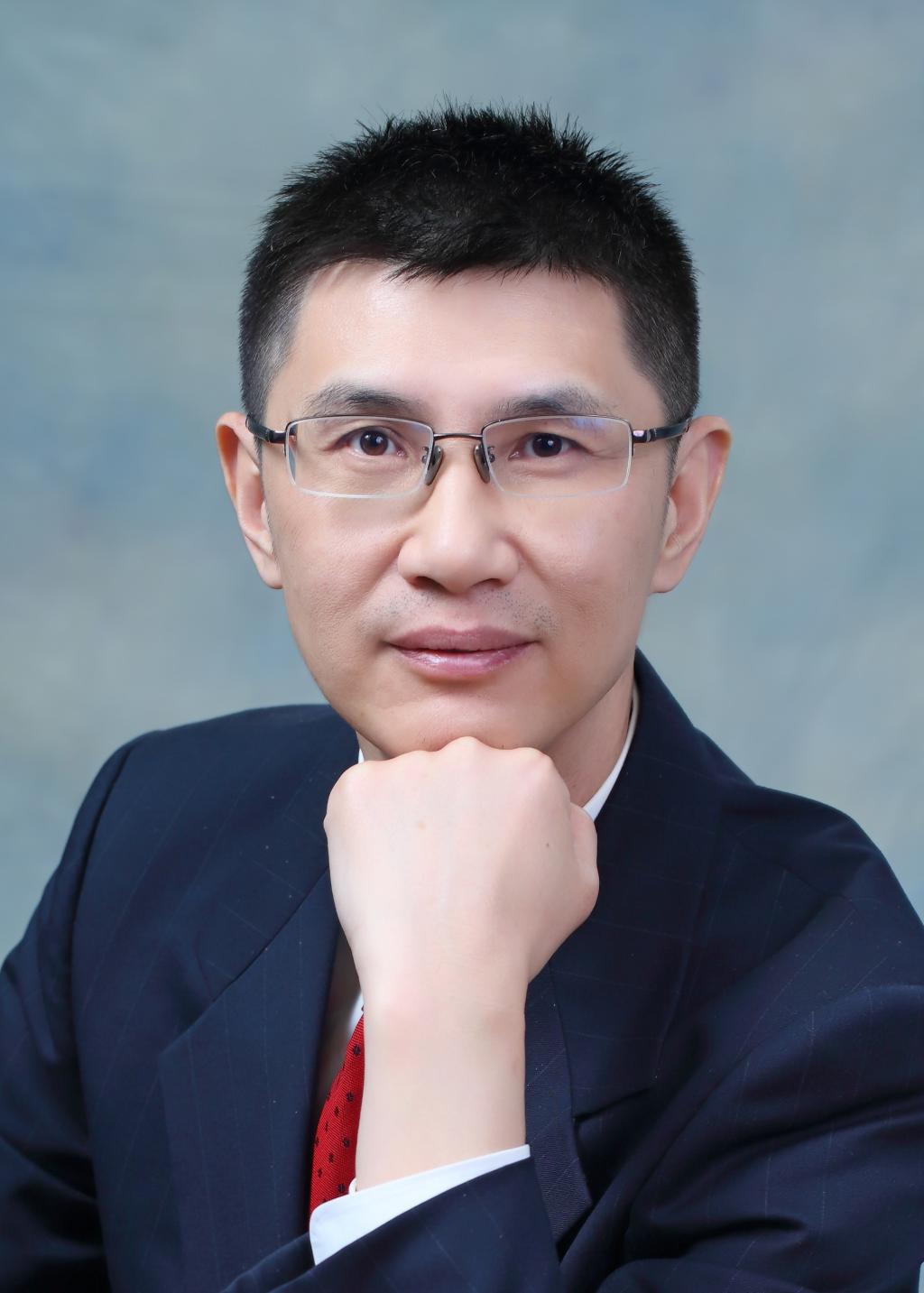}}]
{Tongyu Zhu} received the B.S. degree from Tsinghua University in 1992, and the M.S. degree from Beihang University in 1999. He is currently an associate professor with the State Key Laboratory of Software Development Environment in School of Computer Science, Beihang University, Beijing, China. His research interests include intelligent traffic information processing and network application.
\end{IEEEbiography}

\begin{IEEEbiography}[{\includegraphics[width=1in,height=1.25in,clip,keepaspectratio]{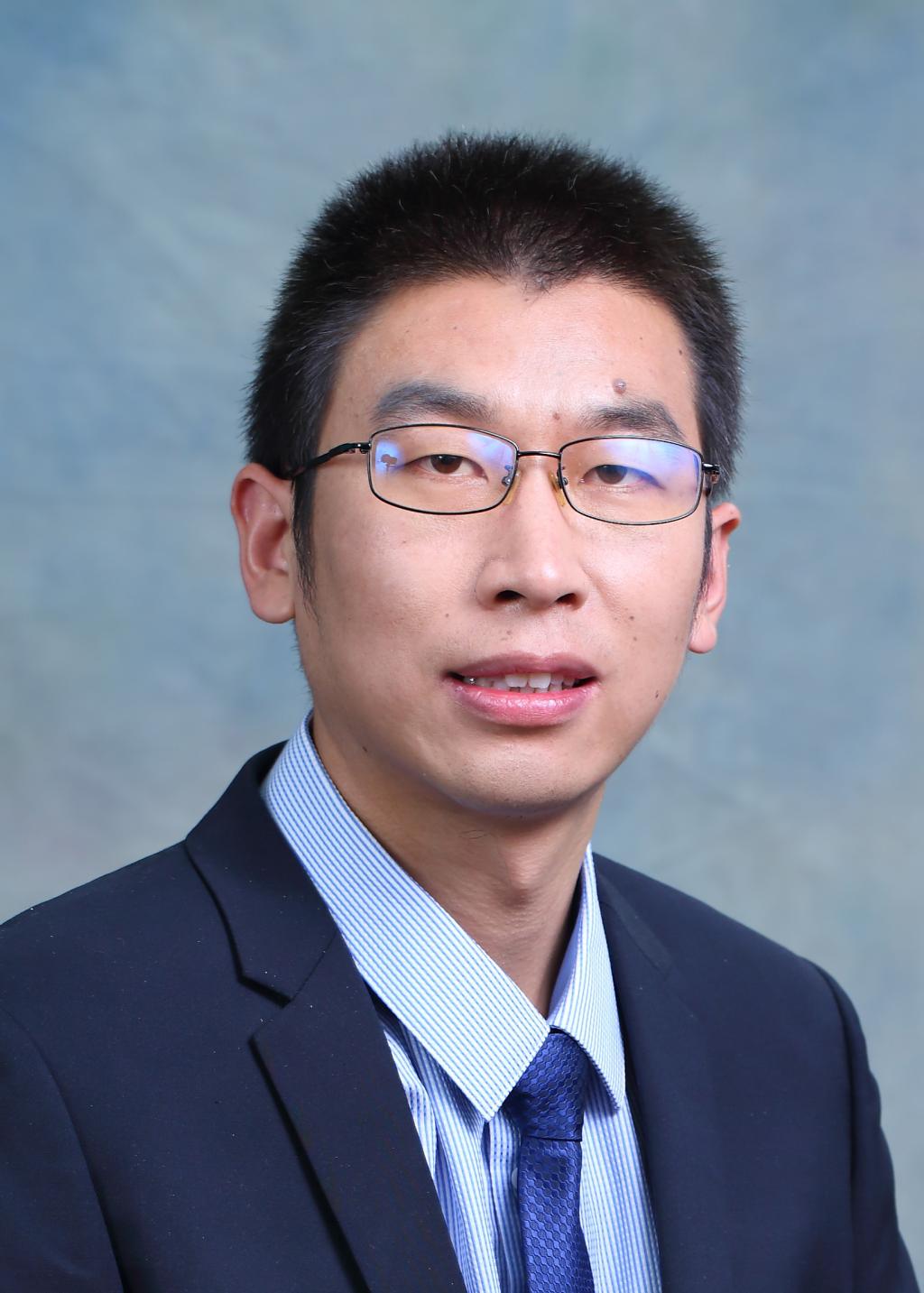}}]
{Leilei Sun} received the B.S., M.S. and Ph.D. degree from Dalian University of Technology in 2009, 2012, and 2017 respectively. He is currently an associate professor with the State Key Laboratory of Software Development Environment (SKLSDE) at School of Computer Science, Beihang University. He was a postdoctoral research fellow from 2017 to 2019 at Tsinghua University. His research interests include machine learning and data mining. 
\end{IEEEbiography}

\vspace{11pt}

\end{document}